\documentclass[lettersize,journal]{IEEEtran}
\usepackage{amsmath,amsfonts}
\usepackage{algorithm}
\usepackage{algorithmic}
\usepackage{array}
\usepackage[caption=false,font=normalsize,labelfont=sf,textfont=sf]{subfig}
\usepackage{textcomp}
\usepackage{stfloats}
\usepackage{url}
\usepackage{verbatim}
\usepackage{graphicx}
\usepackage{multirow}
\usepackage[export]{adjustbox} 
\usepackage{amssymb}
\usepackage{booktabs}
\usepackage{tabularx}
\usepackage{enumitem} 
\usepackage{soul} 
\usepackage[table]{xcolor}
\usepackage[pagebackref,breaklinks,colorlinks]{hyperref}
\usepackage{tablefootnote}

\usepackage[capitalize]{cleveref}
\crefname{section}{Sec.}{Secs.}
\crefname{section}{Section}{Sections}
\crefname{table}{Table}{Tables}
\crefname{table}{Tab.}{Tabs.}

\newcommand{\secref}[1]{Sec.\,\ref{#1}}

\def\BibTeX{{\rm B\kern-.05em{\sc i\kern-.025em b}\kern-.08em
    T\kern-.1667em\lower.7ex\hbox{E}\kern-.125emX}}
\usepackage{balance}
\begin{document}
\title{MVS-TTA: Test-Time Adaptation for Multi-View Stereo via Meta-Auxiliary Learning}
\author{Hannuo Zhang, Zhixiang Chi, Yang Wang, and Xinxin Zuo
\thanks{Hannuo Zhang, Yang Wang, Xinxin Zuo are with Gina Cody School of Engineering and Computer Science, Concordia University,  Montreal, Quebec, Canada
H3G 1M8}
\thanks{Zhixiang Chi is with The Edward S. Rogers Sr. Department of Electrical and Computer Engineering, University of Toronto, Toronto, Ontario, Canada
M5S 3G4}
}


\maketitle

\begin{abstract}
\noindent Recent learning-based multi-view stereo (MVS) methods are data-driven and have achieved remarkable progress due to large-scale training data and advanced architectures. However, their generalization remains sub-optimal due to fixed model parameters trained on limited training data distributions. In contrast, optimization-based methods enable scene-specific adaptation but lack scalability and require costly per-scene optimization. In this paper, we propose MVS-TTA, an efficient test-time adaptation (TTA) framework that enhances the adaptability of learning-based MVS methods by bridging these two paradigms. Specifically, MVS-TTA employs a self-supervised, cross-view consistency loss as an auxiliary task to guide inference-time adaptation. We introduce a meta-auxiliary learning strategy to train the model to benefit from auxiliary-task-based updates explicitly. Our framework is model-agnostic and can be applied to a wide range of MVS methods with minimal architectural changes. Extensive experiments on standard datasets (DTU, BlendedMVS) and a challenging cross-dataset generalization setting demonstrate that MVS-TTA consistently improves performance, even when applied to state-of-the-art MVS models. To our knowledge, this is the first attempt to integrate optimization-based test-time adaptation into learning-based MVS using meta-learning. The code will be available at \url{https://github.com/mart87987-svg/MVS-TTA}.
\end{abstract}
\begin{IEEEkeywords}
Deep Learning for Visual Perception, Computer Vision for Automation, Visual Learning, Test-Time Adaptation, Meta-Learning.
\end{IEEEkeywords}
\section{Introduction}
\label{sec:intro}
\noindent \IEEEPARstart{G}{iven} a set of images from different viewpoints, vision-based 3D reconstruction aims to estimate the 3D structure of a scene or an object~\cite{MVS-Survey, depth-est-survey}. Multi-view stereo (MVS) recovers depth from calibrated, overlapping images to produce dense 3D representations. Recent MVS methods are predominantly learning-based~\cite{mvsformer++, mvsanywhere, geomvsnet}. Despite various implementation strategies, they typically follow a 3-stage pipeline: feature extraction, cost volume construction, and depth estimation via regularization and regression. Recent advances in MVS have demonstrated that scaling up model capacity and training data can significantly improve generalization performance~\cite{mvsformer++, mvsanywhere}. However, data-driven MVS still faces several limitations. In particular, its performance is strongly dependent on the quality and diversity of the training set, and under constrained training computational resources, its generalization ability is limited. On the other hand, during inference, the model parameters are frozen, and the same static model is used to predict 3D scene representation for all scenes. Their adaptability towards specific scenes is limited, hindering their optimal performance.  
\begin{figure}[t]
\scriptsize  
\renewcommand{\arraystretch}{0.1} 
\setlength{\tabcolsep}{1pt}     
\centering
\begin{tabular}{cccc}
        \rotatebox{90}{\parbox{1.9cm}{\centering \textbf{Reference}\\\textbf{Image}}} &
        \includegraphics[height=1.9cm,valign=b]{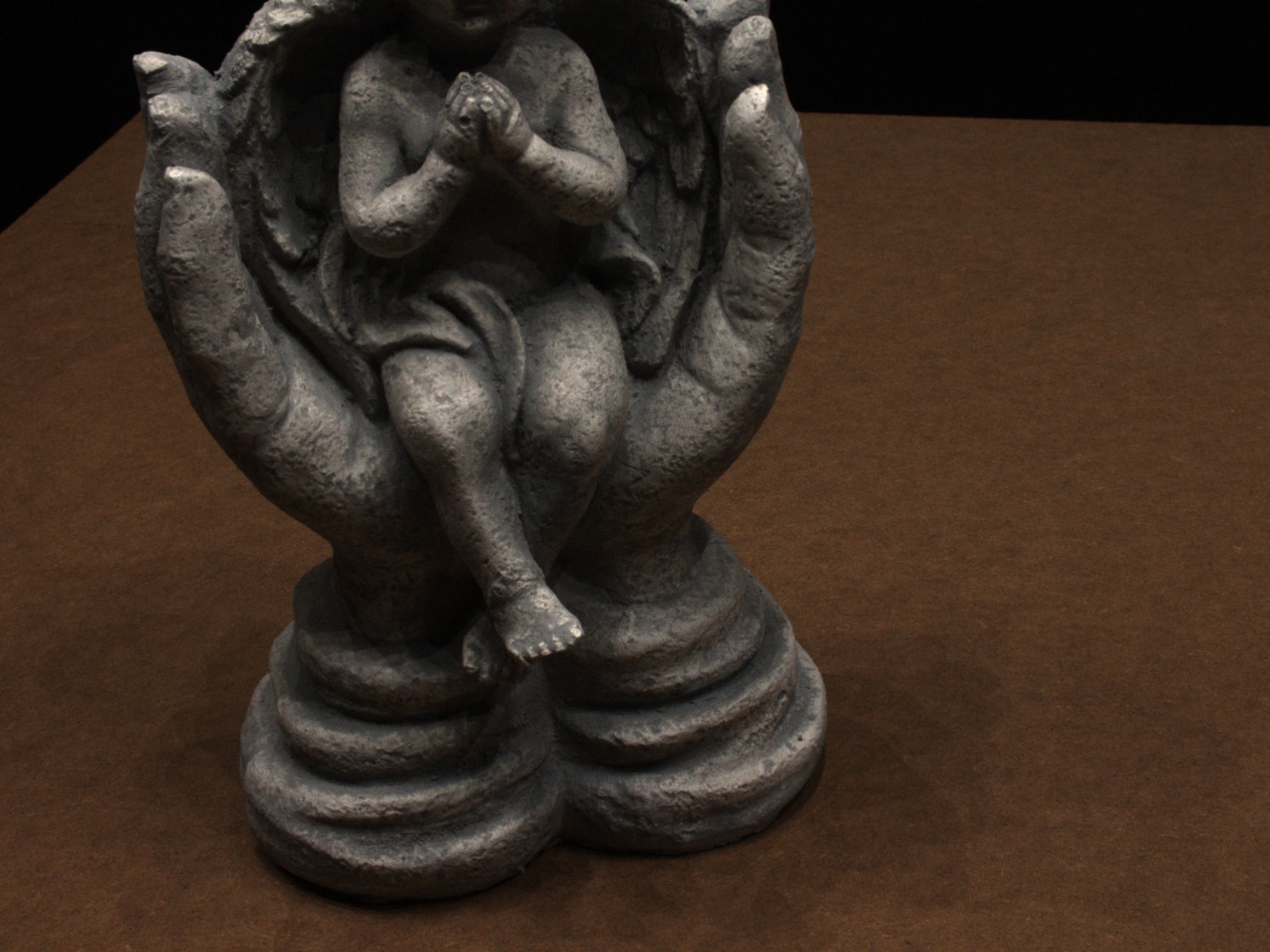} &
        \includegraphics[height=1.9cm,valign=b]{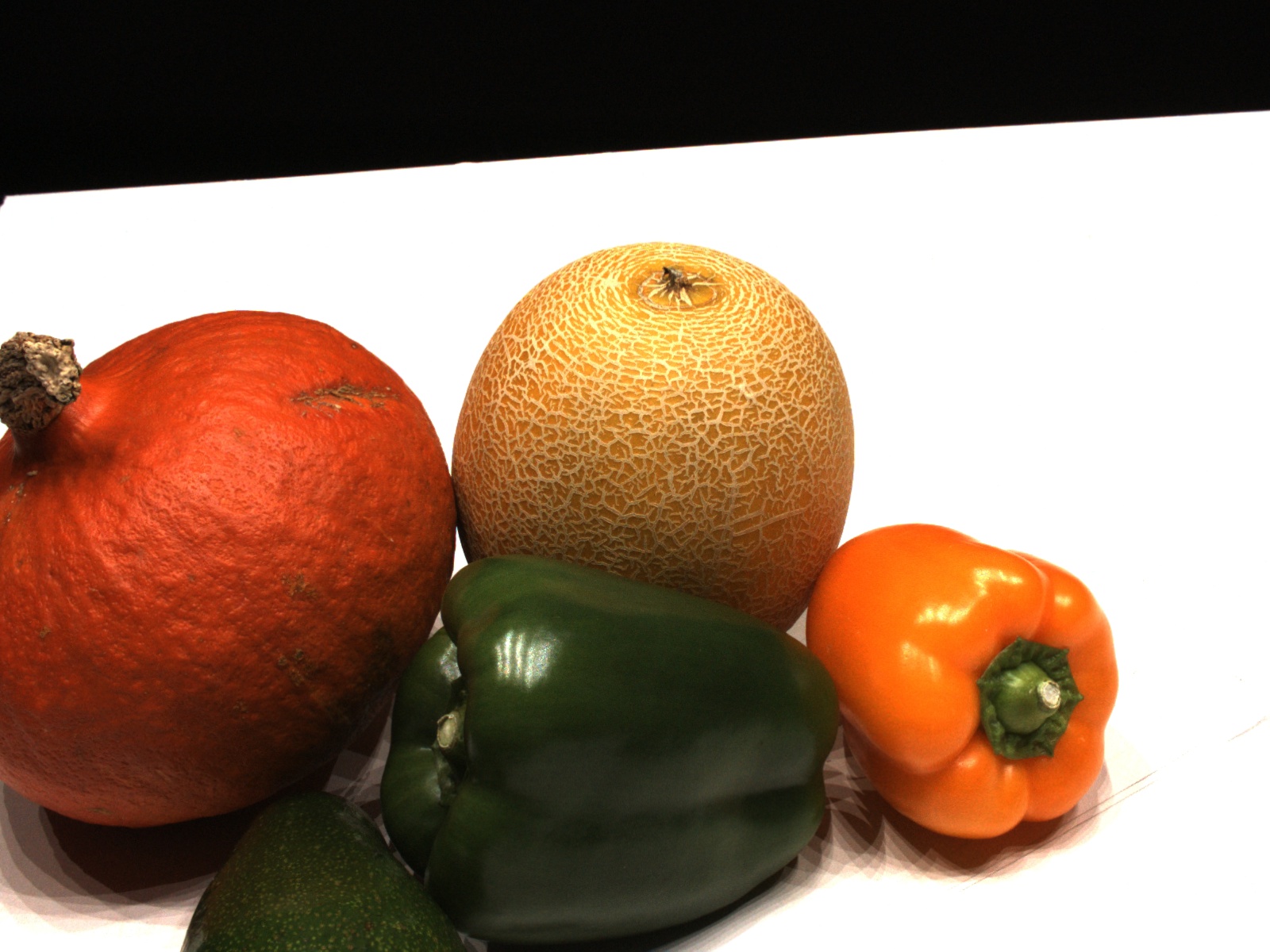} &
        \includegraphics[height=1.9cm,valign=b]{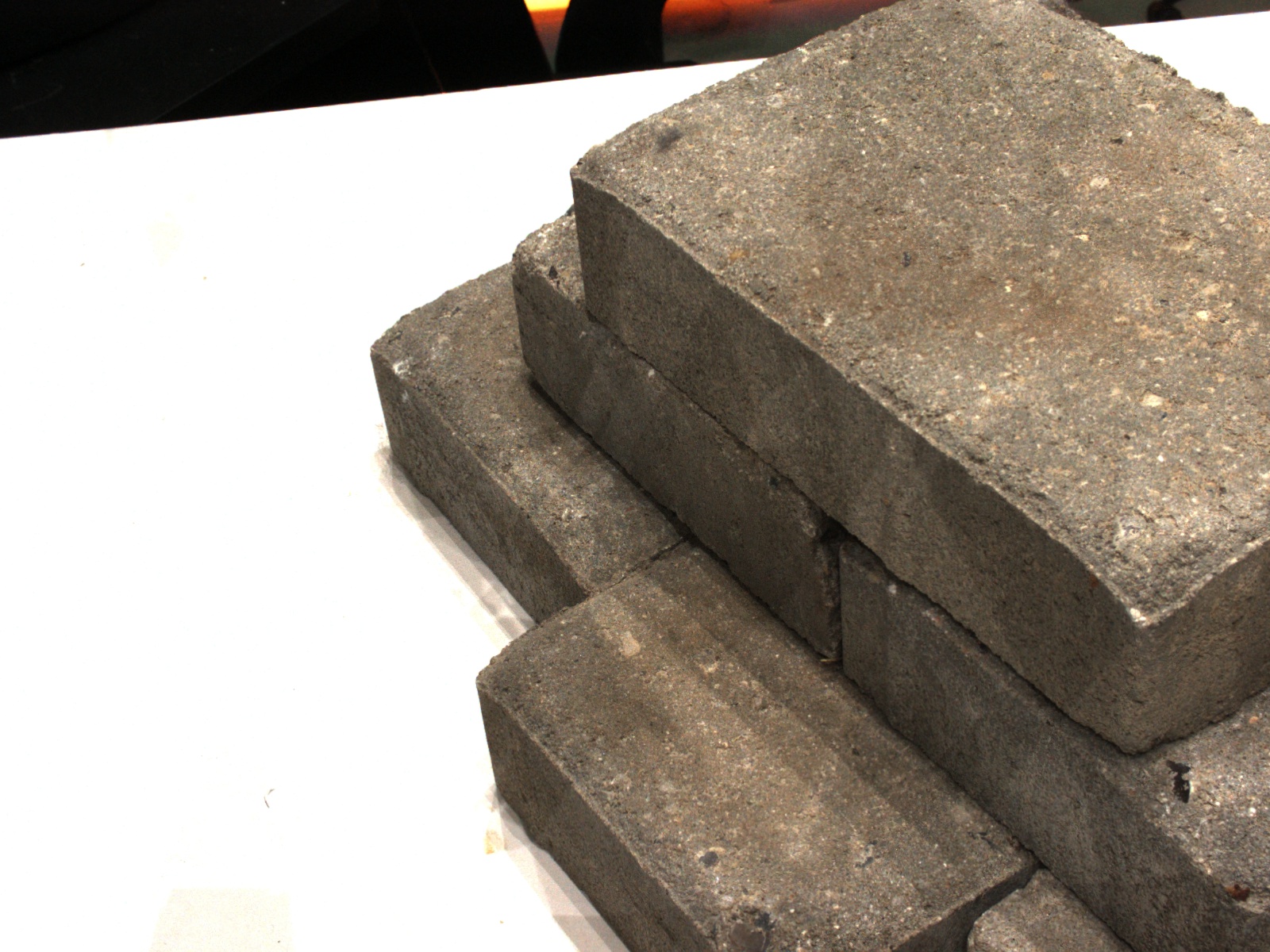} \\
        \rotatebox{90}{\parbox{1.9cm}{\centering \textbf{Ground Truth}\\\textbf{Depth}}} &
        \includegraphics[height=1.9cm,valign=b]{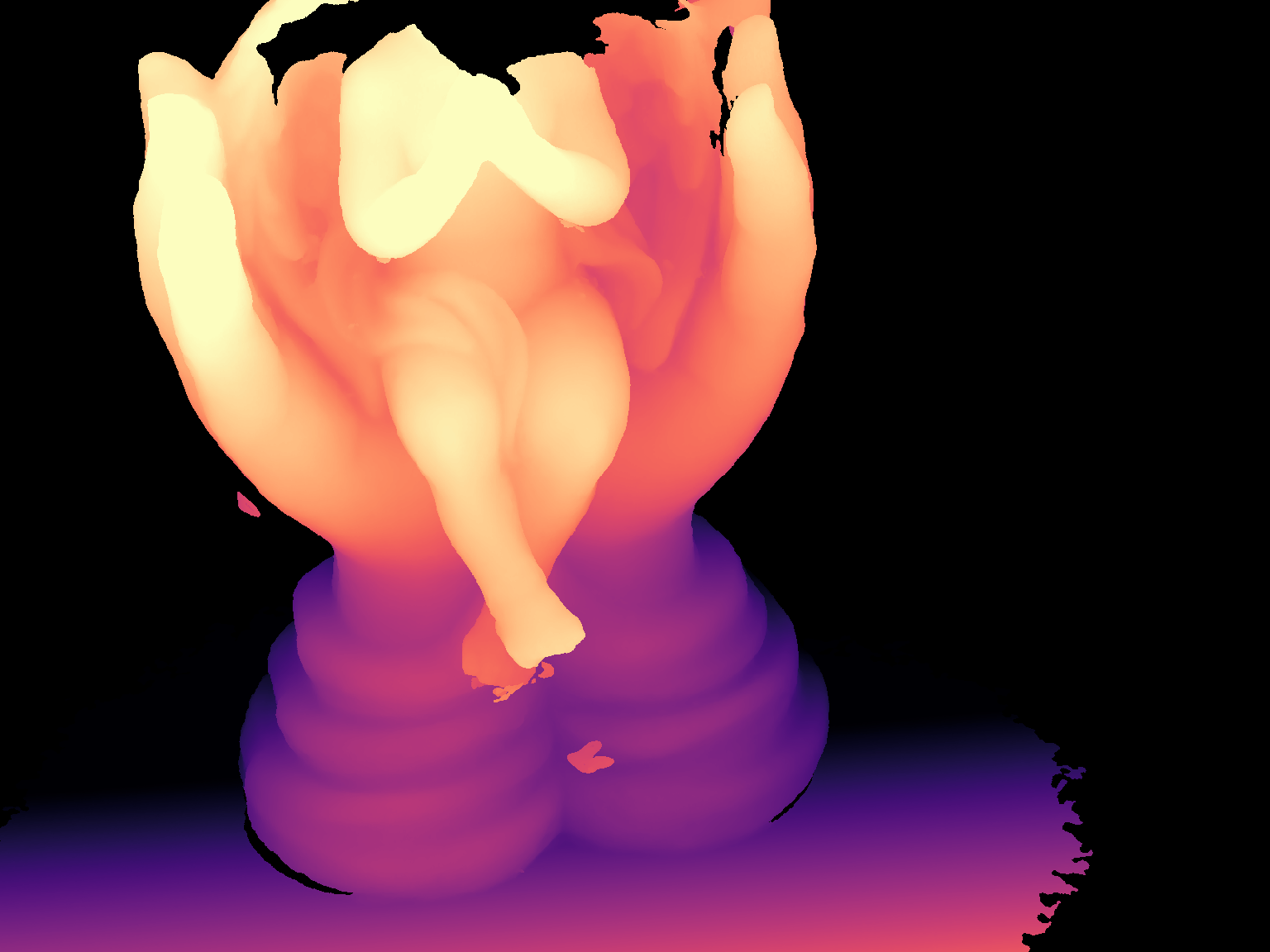} &
        \includegraphics[height=1.9cm,valign=b]{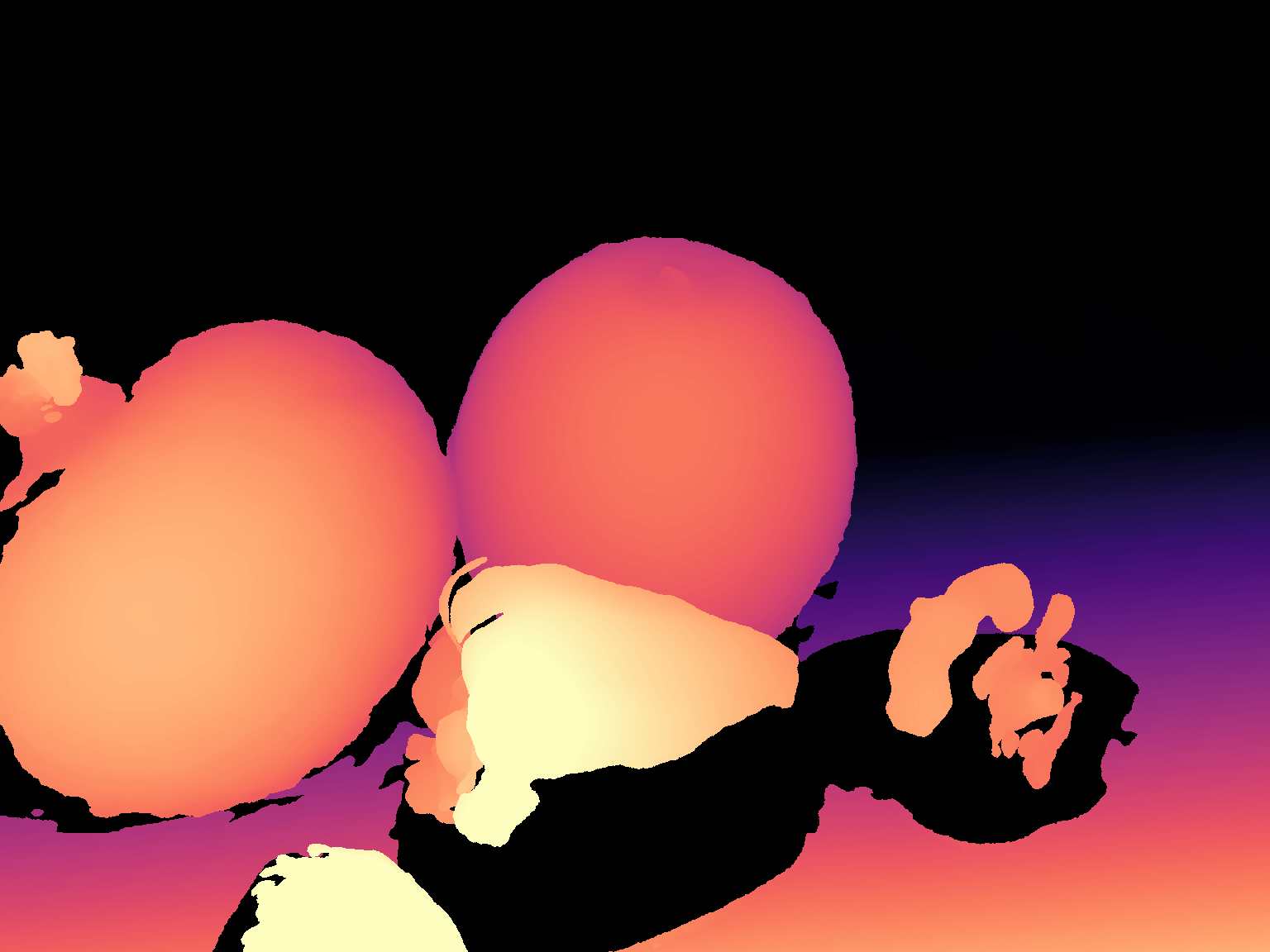} &
        \includegraphics[height=1.9cm,valign=b]{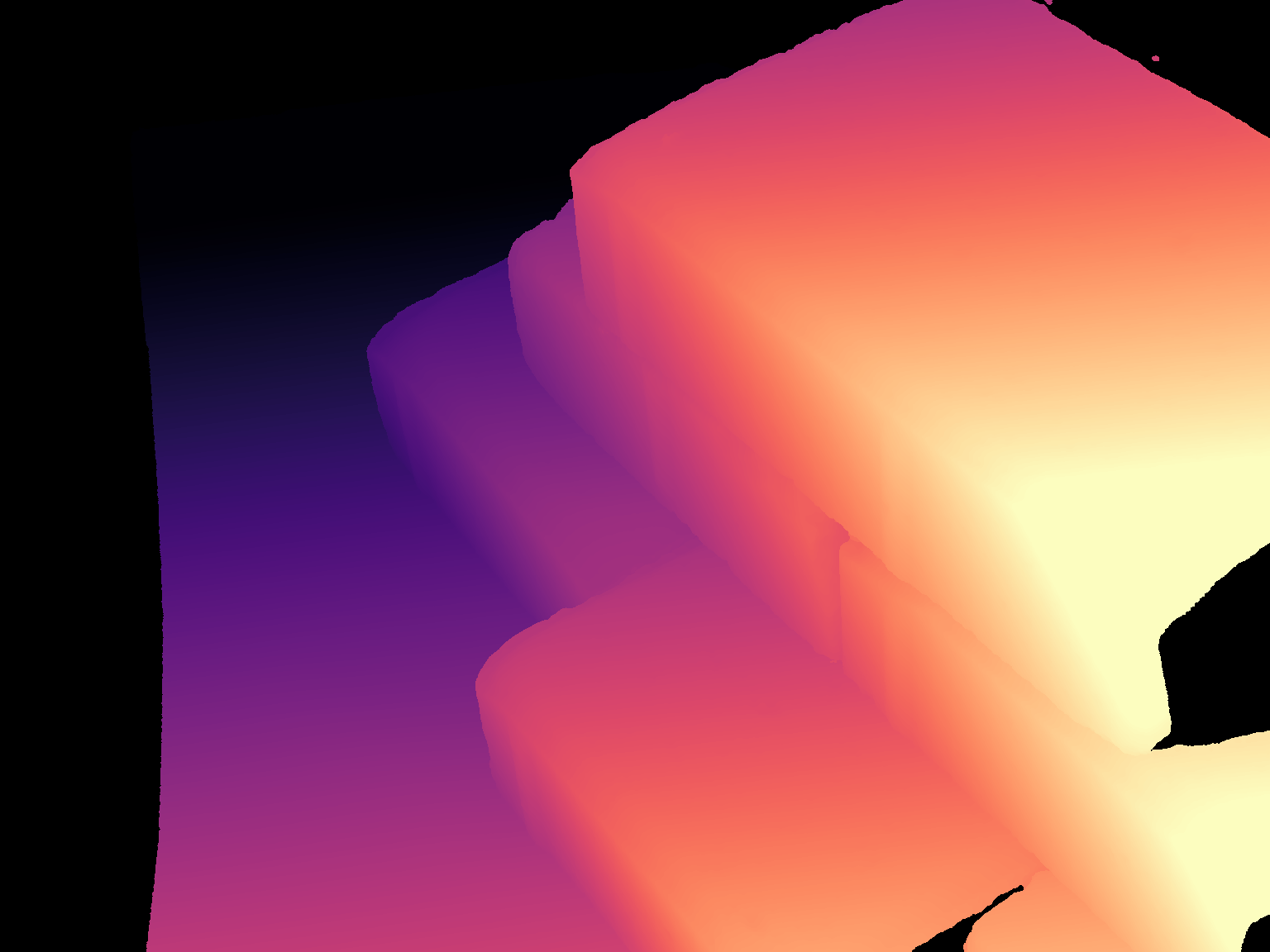} \\
        \rotatebox{90}{\parbox{1.9cm}{\centering \textbf{MVSFormer++}\\\textbf{(Baseline)}}} &
        \includegraphics[height=1.9cm,valign=b]{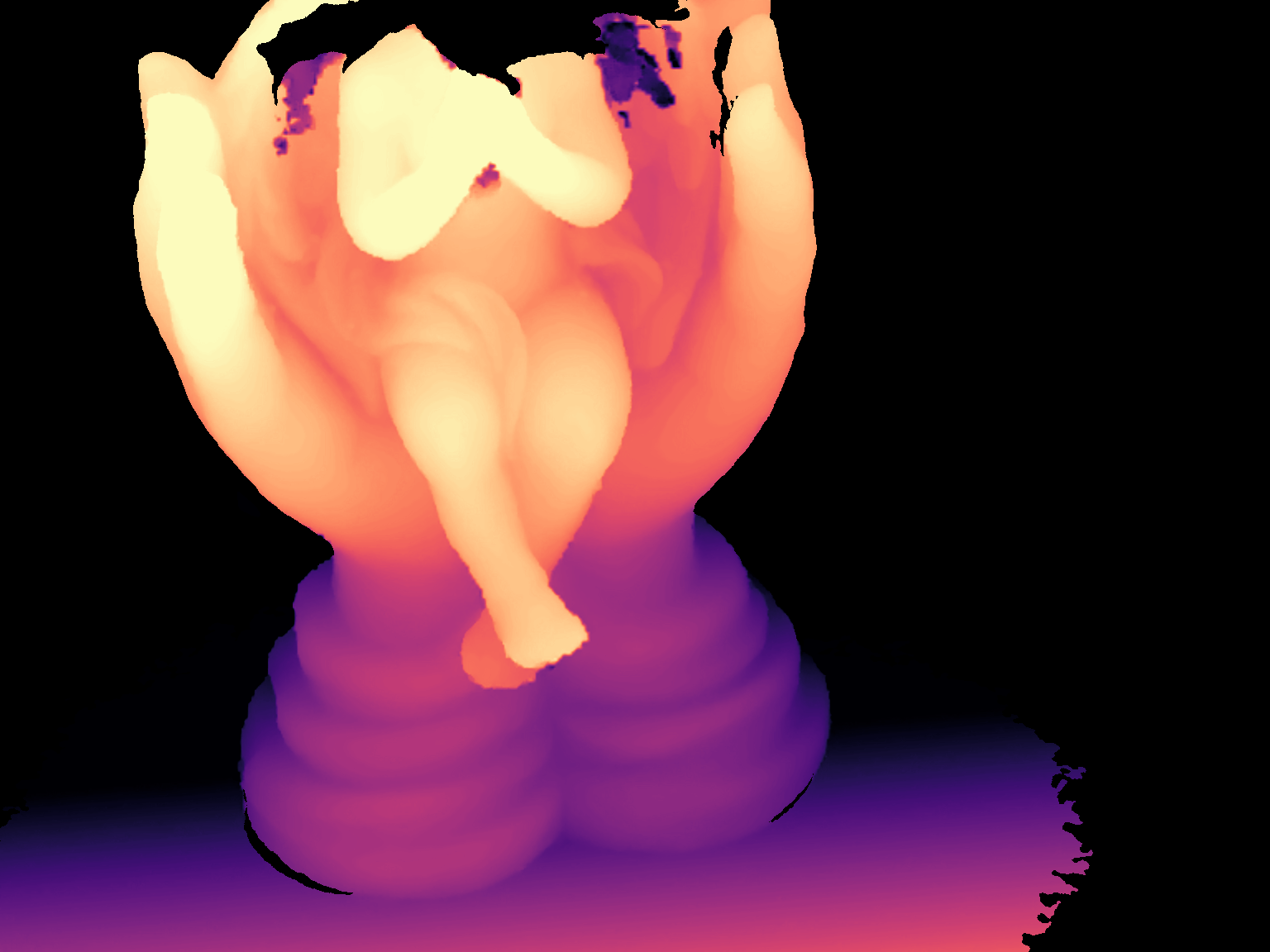} &
        \includegraphics[height=1.9cm,valign=b]{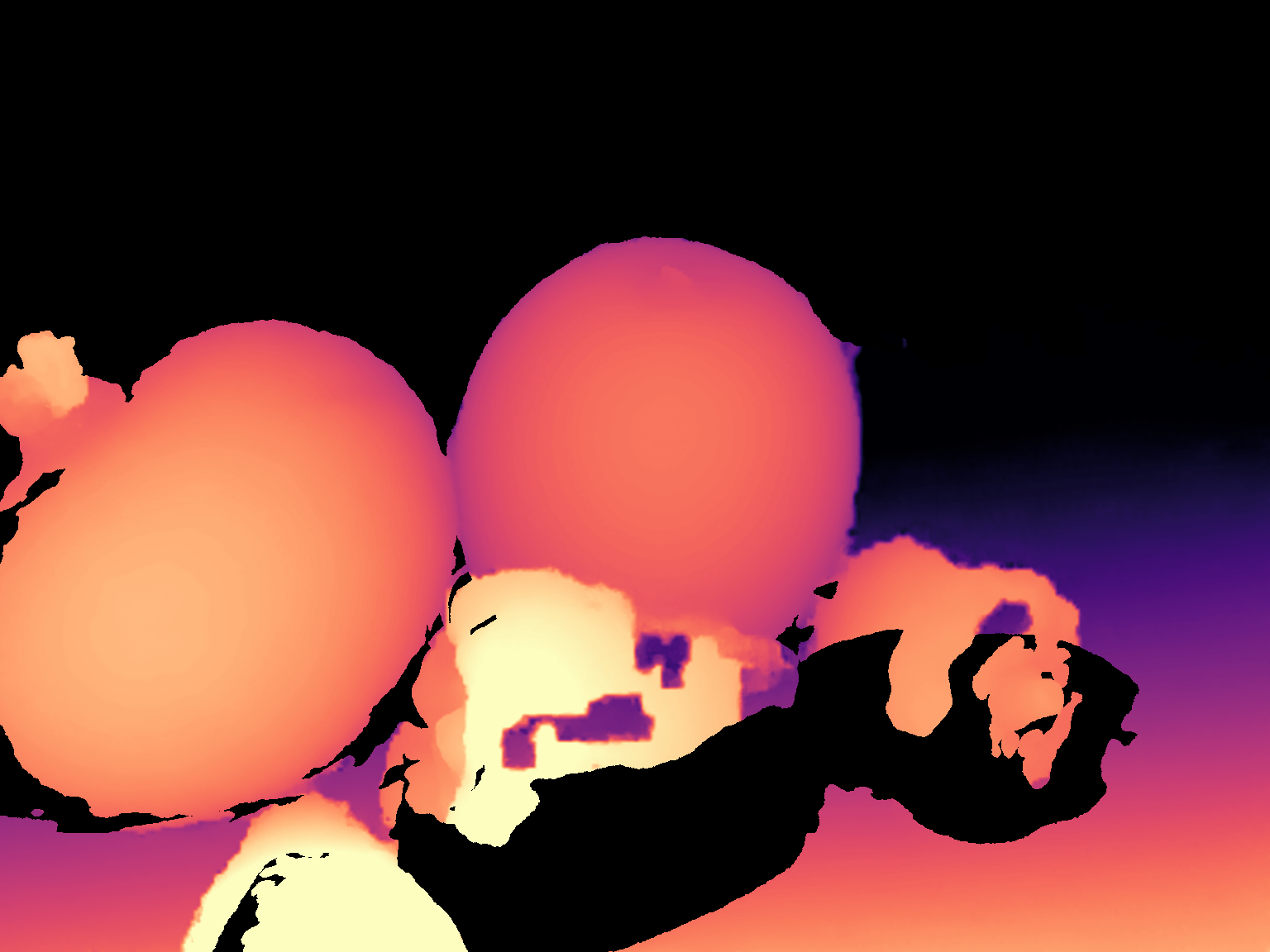} &
        \includegraphics[height=1.9cm,valign=b]{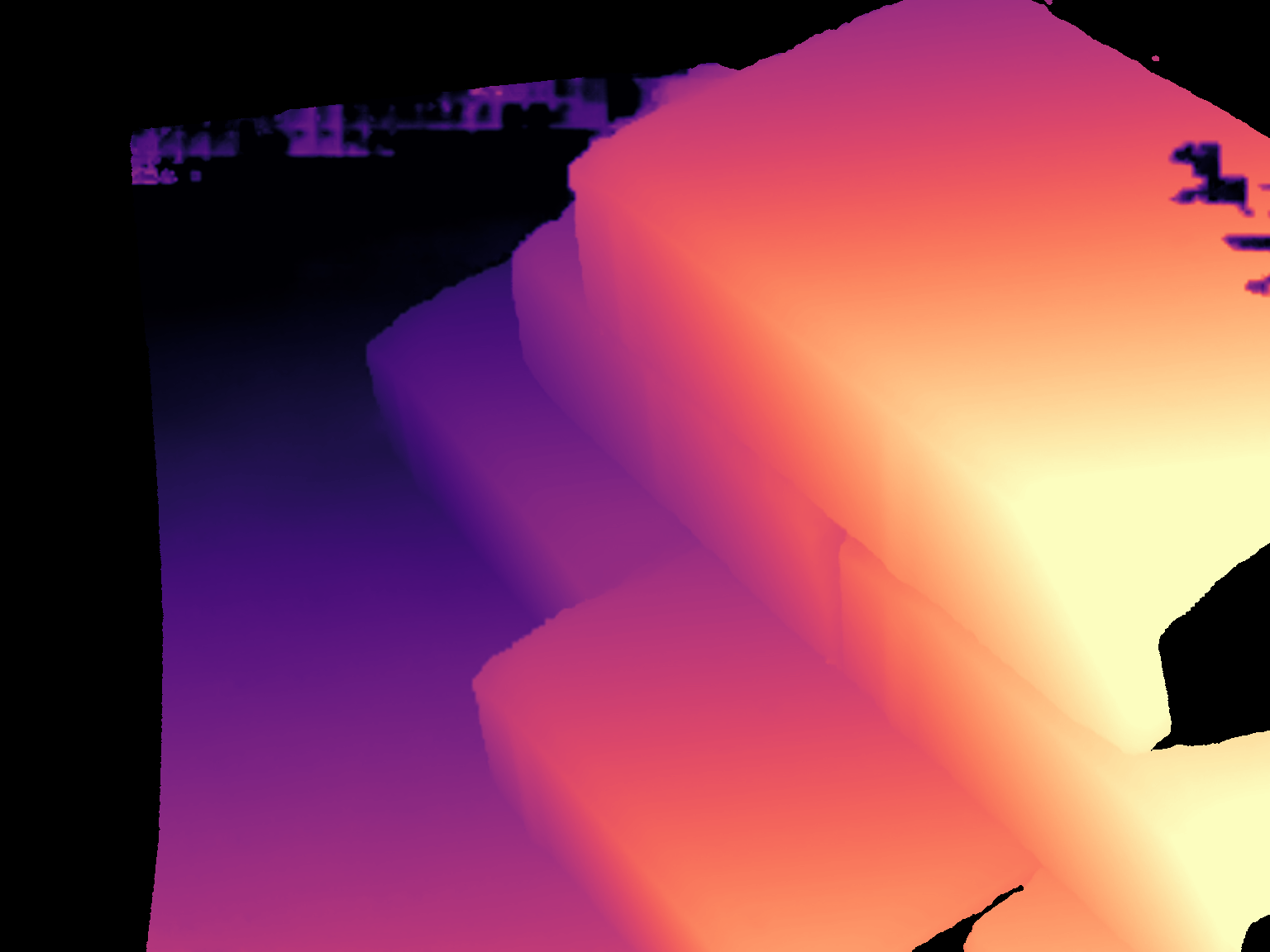} \\
        \rotatebox{90}{\parbox{1.9cm}{\centering \textbf{MVSFormer++}\\\textbf{(MVS-TTA)}}} &
        \includegraphics[height=1.9cm,valign=b]{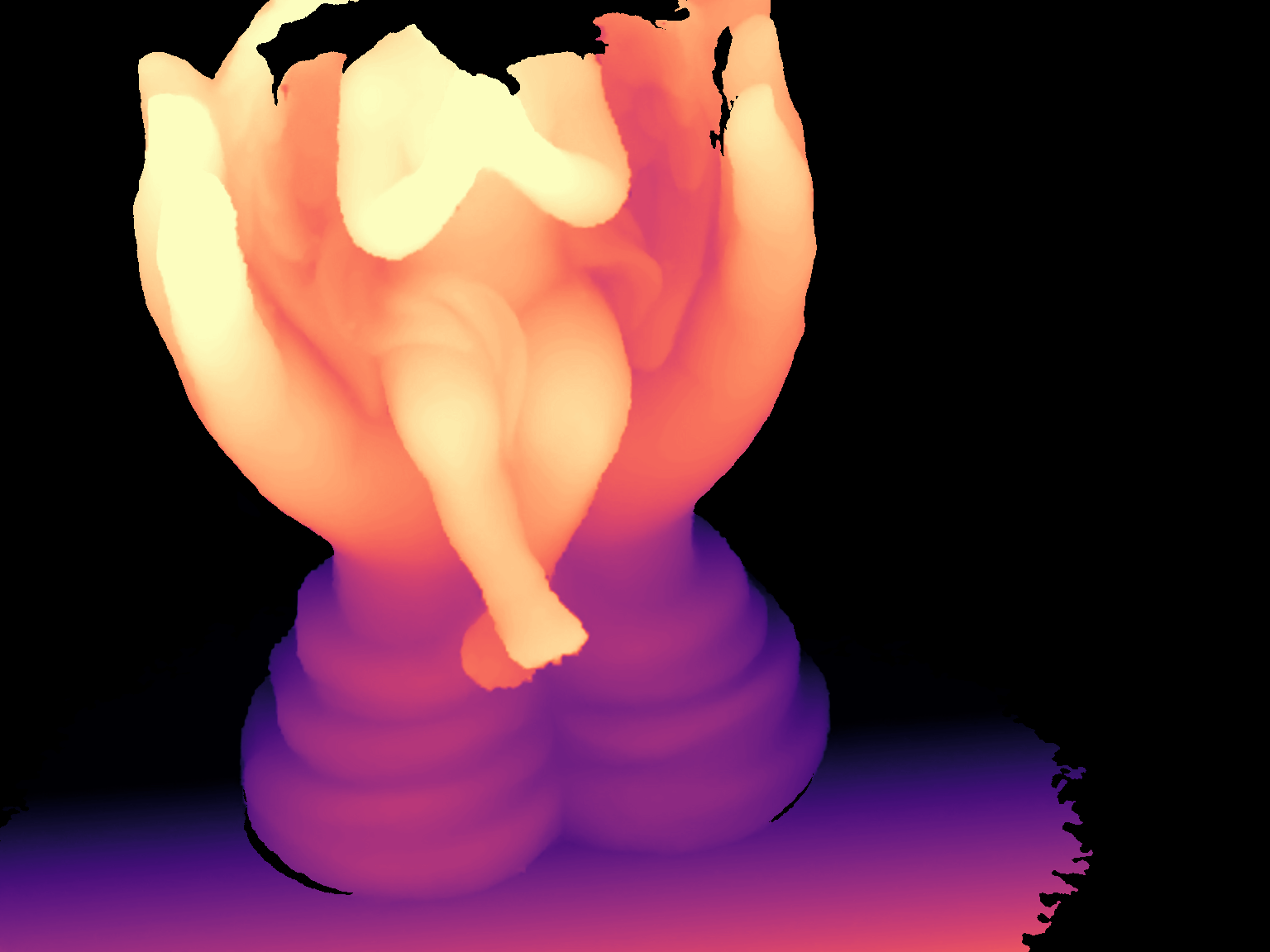} &
        \includegraphics[height=1.9cm,valign=b]{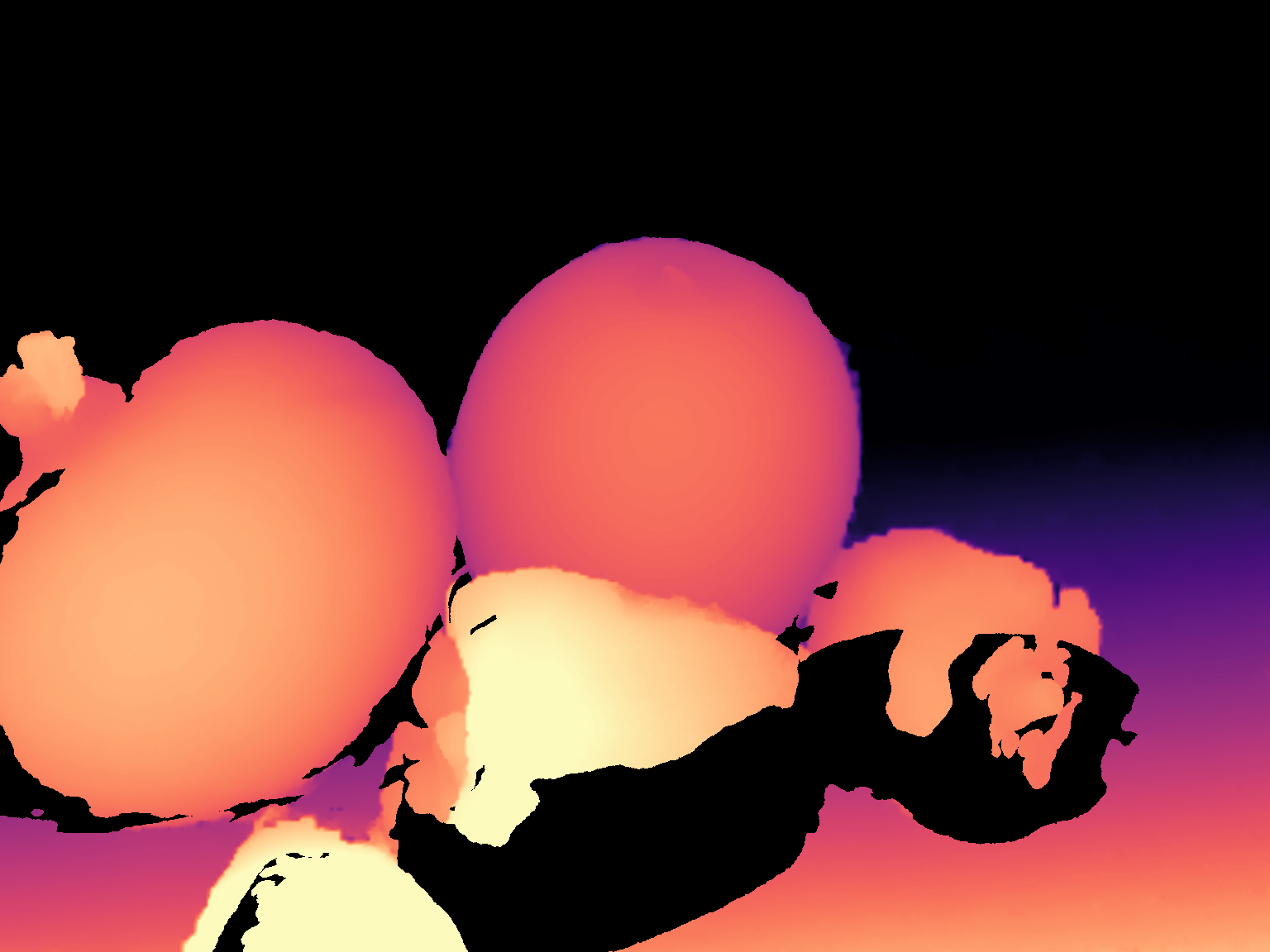} &
        \includegraphics[height=1.9cm,valign=b]{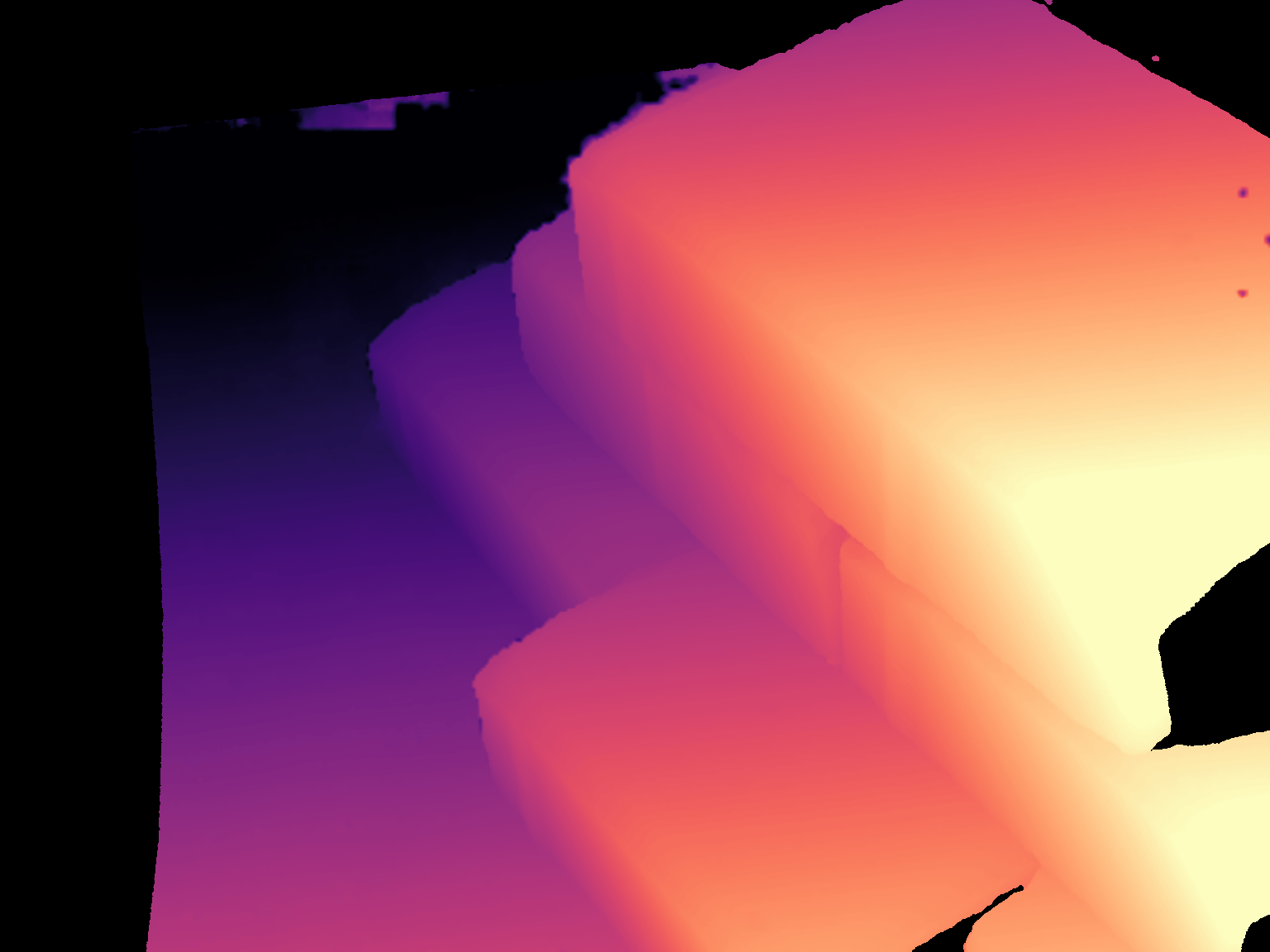} \\
\end{tabular}
\caption{Qualitative comparison on 3 textureless samples from the DTU dataset. From top to bottom: reference image, ground-truth depth, prediction by MVSFormer++ (baseline), and prediction by MVSFormer++
+ MVS-TTA.}
\label{fig:qualitative-textureless}
\end{figure}
Optimization-based 3D reconstruction, as another line of research, reconstructs a scene after directly optimizing representations for each individual scene. Traditional Structure from Motion (SfM)~\cite{SfM}, Neural Radiance Fields (NeRF)~\cite{nerf}, and 3D Gaussian Splatting~\cite{3DGS} are the representative optimization-based methods. Those works exhibit strong expressive power and high reconstruction fidelity, as they directly optimize scene-specific parameters to fit the observed views. This enables them to capture fine-grained appearance and lighting effects from a given scene, often resulting in superior reconstruction accuracy compared to learning-based approaches. However, optimization-based methods lack the priors from the large-scale dataset and cannot directly generalize to unseen scenes. Furthermore, they normally suffer from high computational cost, as they require time-consuming per-scene optimization. Therefore, their scalability and applicability in real-time or large-scale multi-scene scenarios are limited.

In this work, we propose a \textbf{t}est-\textbf{t}ime \textbf{a}daptation~\cite{TTA-Survey} framework designed explicitly for learning-based MVS~\cite{MVS-Survey} (namely \textbf{MVS-TTA}), enabling scene-specific adaptation at inference. MVS-TTA combines the strengths of both learning-based (efficiency and scalability) and optimization-based approaches (adaptability) while addressing their respective limitations. Specifically, MVS-TTA bridges this gap by enabling a pre-trained MVS model to perform lightweight, scene-specific optimization at test time without training from scratch. To achieve this, we draw inspiration from self-supervised MVS methods~\cite{self-mvs, robust-photo} and adopt a cross-view photometric consistency loss as the optimization objective during test-time. This loss enforces pixel-wise alignment between the reference image and source-view reconstructions, combining color and structural similarity metrics to handle occlusions and illumination changes. To further enhance test-time adaptability, we propose meta-auxiliary learning~\cite{A-MAML-deblur, A-MAML-future-depth, A-MAML-point,wu2023metagcd} based on the model-agnostic meta-learning (MAML)~\cite{MAML} principle. This meta-learning strategy explicitly trains the model to benefit from test-time adaptation by encouraging it to rapidly improve performance on the primary task with a few gradient steps on an auxiliary task. Specifically, we define the primary task as supervised depth inference and the auxiliary task as unsupervised cross-view consistency. Meta-auxiliary training backpropagates through the adaptation process, optimizing model parameters to be well-suited for fast and effective test-time adaptation.

Notably, the proposed MVS-TTA is model-agnostic and can be applied to a wide range of learning-based MVS methods that differ significantly in architecture, parameter count, and depth inference strategies. To the best of our knowledge, this is the first attempt to integrate optimization-based adaptation into learning-based MVS via meta-auxiliary learning. Comprehensive experiments and ablation studies demonstrate the effectiveness and general applicability of MVS-TTA in enhancing the state-of-the-art methods for MVS. Our key contributions are as follows:
\begin{itemize}[label=\textbullet]
\item We propose a novel test-time adaptation framework for learning-based MVS. It enables MVS models to adapt to novel scenes during inference using a self-supervised objective.
\item We introduce meta-auxiliary learning to explicitly optimize MVS models, enabling them to benefit from auxiliary-task-based adaptation at inference and leading to more effective scene-specific adaptation.
\item We demonstrate the broad applicability and consistent effectiveness of MVS-TTA across representative MVS methods, standard datasets, and a challenging cross-dataset generalization setting. Extensive experiments demonstrate that MVS-TTA yields performance improvements when applied to various MVS models.
\end{itemize}
\section{Related Works}
\subsection{Learning-based Multi-view Stereo}
\noindent MVS predicts dense 3D representations from multiple calibrated views and is widely applied in remote sensing~\cite{MVS_application_RS}, robotic vision~\cite{MVS_application_robotics}, and autonomous driving~\cite{MVS_application_driving}. MVSNet~\cite{mvsnet} builds a cost volume via homography warping and estimates depth with 3D CNNs. To improve scalability, CasMVSNet~\cite{cascade} improves scalability through a coarse-to-fine cascade of cost volumes. TransMVSNet~\cite{transmvsnet} employs feature matching transformers to capture long-range correspondences. MVSFormer~\cite{mvsformer} and MVSFormer++~\cite{mvsformer++} further explore the use of pre-trained Vision Transformers (ViTs) for MVS~\cite{vit, chu2021twins, dinov2}, and adopt an efficient multi-scale training strategy to improve generalization across different resolutions. Self-supervised MVS methods~\cite{robust-photo,self-mvs} leverage photometric consistency loss, pseudo-label refinement, and flow-depth consistency loss to estimate depth without ground-truth supervision. More recently, MVSAnywhere~\cite{mvsanywhere} improves generalization across diverse depth ranges and scene types. VGGT~\cite{vggt} is a 1.2B-parameter feed-forward 3D foundation model trained on extensive multi-dataset 3D annotations, fundamentally different from our lightweight MVS-TTA setting. Compared to these methods, our proposed MVS-TTA framework improves the generalization performance of diverse MVS architectures without requiring additional 3D annotations or a computationally expensive training process.
\subsection{Meta-auxiliary Learning}
\noindent Meta-learning aims to train models that can quickly adapt to new tasks with limited data by leveraging experience from related tasks. Among gradient-based approaches, Model-Agnostic Meta-Learning~\cite{MAML} is widely used. It optimizes model parameters such that a few gradient updates on a new task lead to fast and effective adaptation. Building on this idea, Chi et al.~\cite{A-MAML-deblur} propose a meta-auxiliary learning scheme that leverages a self-supervised auxiliary task for efficient test-time adaptation, improving robustness under distribution shifts. Liu et al.~\cite{A-MAML-future-depth} extend meta-auxiliary learning to future depth prediction, jointly training a depth estimation and image reconstruction branch. The self-supervised auxiliary task enables test-time adaptation, improving generalization to unseen video sequences. Hatem et al.~\cite{A-MAML-point} propose Point-TTA, which uses meta-auxiliary learning with three self-supervised tasks to enable test-time adaptation for point cloud registration, improving generalization to unseen 3D data. Tonioni et al.~\cite{tonioni2019learning} introduce a meta-learning framework for stereo matching that continuously adapts to video streams using an unsupervised reprojection loss with a confidence weighting network. While effective in stereo settings, this approach is limited to simple two-view baselines and may suffer from training–testing mismatch and catastrophic forgetting. MVS-TTA employs the meta-auxiliary learning strategy to enable MVS models to improve their performance through a few gradient update steps based on a self-supervised loss.
\section{Method}
\begin{figure*}[t]
    \centering
    \includegraphics[width=\textwidth]{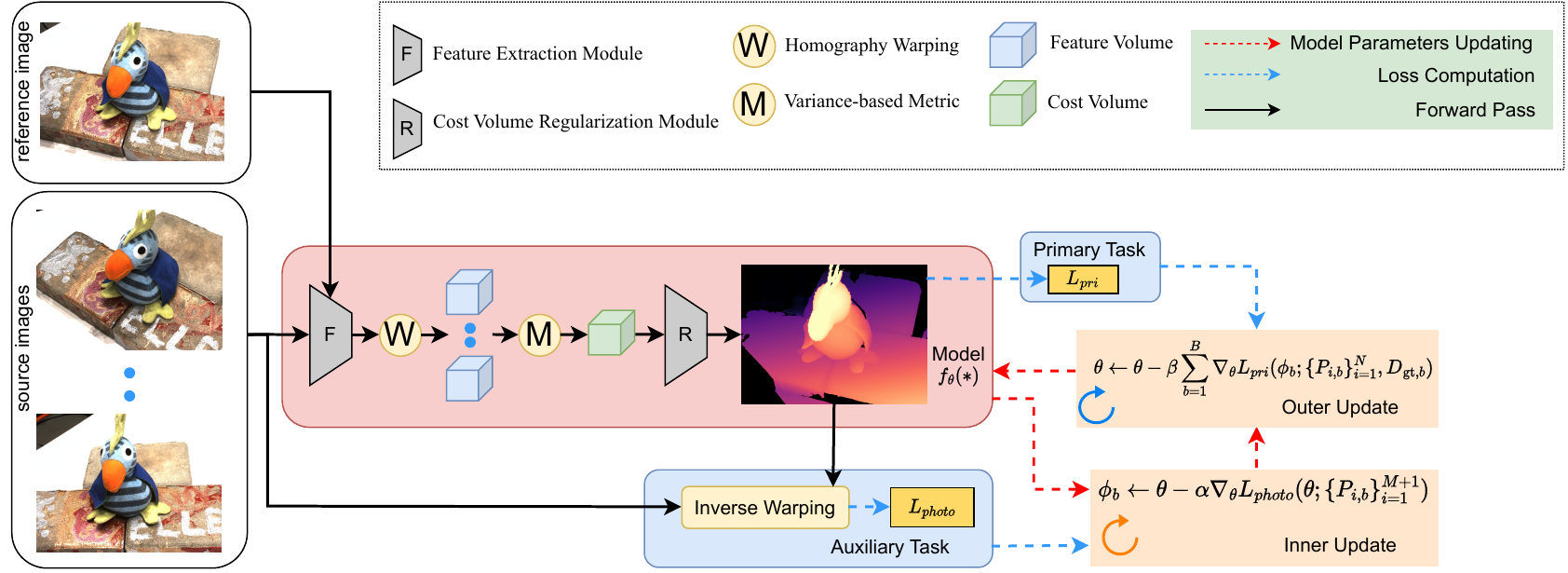}
    \caption{Overview of meta-auxiliary training in the proposed MVS-TTA framework. Given a batch of samples $\{ \{ {P_{i,b}}\} _{i = 1}^{M + 1},{D_{{\rm{gt}},b}}\} _{b = 1}^B$, the meta training process follows a nested loop structure. For each sample $(\{ {P_{i,b}}\} _{i = 1}^{M + 1},{D_{{\rm{gt}},b}})$, we first adapt the model parameters $\theta $ for a few steps using the photometric consistency loss ${L_{photo}}$ as an auxiliary task. Then, in the outer loop, the adapted model ${\phi _b}$ performs the primary task of depth inference, where the primary loss ${L_{pri}}$ measuring the discrepancy between the predicted depth map and the ground-truth annotation is computed and used to update the original model parameters $\theta $.}
    \label{fig:method_overview}
\end{figure*}
\begin{figure}[t]
    \centering
    \includegraphics[width=\linewidth]{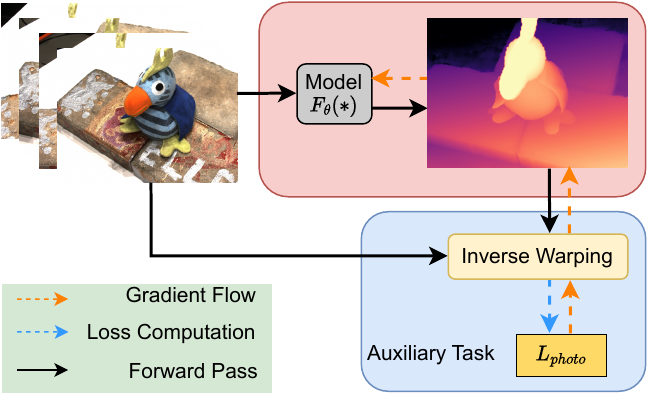}
    \caption{Overview of the test-time adaptation procedure. Given a test sample, we adapt the meta-trained model for a few steps using the auxiliary task of photometric consistency loss. The adapted model is then used to infer the depth map of that image.}
    \label{fig:tto_process}
\end{figure}
\noindent \textbf{Overview:} Given a set of $N$ overlapping images $\{ {I_i}\} _{i = 1}^N$ with known camera intrinsics $\{ {K_i}\} _{i = 1}^N$ and extrinsics $\{ {R_i},{t_i}\} _{i = 1}^N$, the objective of learning-based MVS~\cite{mvsnet} is to estimate a depth map ${D_{{\rm{pred}}}}$ for a designated reference view ${I_1}$. For notational clarity, we define each posed image as ${P_i} = \{ {I_i},{K_i},{R_i},{t_i}\} $, and denote the set of $N$ posed images as $\{ {P_i}\} _{i = 1}^N$. This task is typically achieved by learning a function parameterized by weights $\theta $ as: ${D_{{\rm{pred}}}} = {f_\theta }(\{ {P_i}\} _{i = 1}^N)$. In our framework, the backbone MVS model is first trained with ground-truth depth maps. To enhance the ability of the model to adapt to novel test-time scenes, we introduce a test-time adaptation mechanism guided by a self-supervised photometric consistency loss~\cite{robust-photo}. Furthermore, to improve the effectiveness of TTA, we propose a meta-auxiliary learning strategy~\cite{A-MAML-future-depth} that optimizes the model to be amenable to lightweight self-supervised adaptation at inference time. In this section, we first describe the standard pipeline for learning-based MVS, then present the formulation of our test-time adaptation objective, followed by the meta-auxiliary training strategy and test-time adaptation procedure.
\subsection{Depth Estimation for Primary Task}
\noindent In this section, we provide a brief overview of the general framework for learning-based MVS methods. To be noted, our framework is model-agnostic and can be applied to a wide range of learning-based MVS architectures. 
First of all, most models follow a general pipeline consisting of three stages: (1) feature extraction from all input views using 2D CNNs~\cite{cascade, geomvsnet, fpn} or transformers~\cite{transmvsnet, mvsformer++, dinov2}, (2) differentiable homography warping to build cost volumes on fronto-parallel depth planes, and (3) depth prediction via cost volume regularization. 

Specifically, homography warping is commonly implemented by projecting each source-view feature map onto a hypothetical depth plane $d$ in the reference view using:
\begin{equation}
H_{i}(d)=K_{i} \cdot R_{i}\cdot (I - \frac{(t_{1}-t_{i})\cdot{n_{1}}^{T}}{d}) \cdot {R_{1}}^{T} \cdot {K_{1}}^{-1},
\label{eq_mvsmapping}
\end{equation}
where ${{n_1}}$ denotes the principal axis of the reference view. $H_i(d)$ is the homography matrix that defines the planar transformation between the $i$-th source view and the reference view at depth $d$. It maps a point $u$ in the reference feature map to its corresponding location $\hat{u}$ in the source view via:
\begin{equation}
\label{eq:coord-mapping}
{\hat u} \sim {H_i}(d) \cdot u,
\end{equation}
where $u$ denotes a pixel coordinate in the reference view. Using the coordinate correspondence defined in Eq. \ref{eq:coord-mapping}, we sample features from the image feature maps $\{ {F_i}\} _{i = 1}^N$( ${F_i} \in {\mathbb{R}^{H \times W \times F}}$ ) via differentiable bilinear sampling to construct the feature volumes $\{ {V_i}\} _{i = 1}^N$ (${V_i} \in {\mathbb{R}^{H \times W \times D \times F}}$), where $H$ and $W$ denote the height and width of the feature maps, $F$ is the feature dimension, and $D$ is the number of depth hypotheses. Feature volumes $\{ {V_i}\} _{i = 1}^N$ are typically aggregated into a cost volume $C$ using a variance-based metric $\mathcal{M}$:
\begin{equation} 
\label{eq:metric}
C = {\cal M}({V_1}, \cdots ,{V_N}) = \frac{{\sum\limits_{i = 1}^N {{{({V_i} - {\rm{ }}{\bar V})}^2}} }}{N},
\end{equation}
where $\bar{V}$ is the mean feature volume. A CNN or transformer-based regularization module is then applied to refine the cost volume and produce a probability volume $Pr$, from which the expected depth is estimated:
\begin{equation}
\label{prob2depth}
D_{\text{pred}} = \sum\limits_{d = {d_{min}}}^{{d_{max}}} d  \times Pr(d).
\end{equation}

We denote the supervised training objective as the primary loss:
\begin{equation}
\label{eq:pri-loss}
{L_{pri}}(\theta ;\{ {P_i}\} _{i = 1}^N,{D_{{\rm{gt}}}}),
\end{equation}
which measures the difference between predicted and ground-truth depth maps. The primary loss follows the original loss function defined by each backbone model. However, for models that predict depth maps using classification over a probability volume~\cite{geomvsnet, transmvsnet}, we replace the original loss with an $L_1$ loss on the predicted depth map.
\subsection{Cross-view Consistency as Auxiliary Task}
\noindent To enable scene-specific adaptation during inference, we employ a self-supervised objective based on cross-view photometric consistency~\cite{robust-photo, self-mvs}. This loss is defined over input calibrated images and does not require ground-truth 3D data, making it well-suited for test-time adaptation.\\
{\bf Cross-view Photometric Consistency} The core idea is that a correctly estimated depth map should yield photometric consistency between the reference view and the warped source views. Given a predicted depth map $D_{\text{pred}}$ for reference image $I_1$ and $M$ calibrated source images $\left\{ {{I_m}} \right\}_{m = 2}^{M + 1}$ (with $M > N - 1$), where $N - 1$ is the number of source views used during primary task training, each source image is warped to the reference view via inverse warping :
\begin{equation}
\label{eq:Homography}
\hat u \sim {H_m}({D_{{\rm{pred}}}}(u)) \cdot u.
\end{equation}
The warped source image $\hat{I}_1^m$ is then obtained through bilinear sampling:
\begin{equation}
\label{eq:Coord-Mapping}
\hat I_1^m(u) = {I_m}(\hat u).
\end{equation}
To improve robustness to occlusions and lighting variations, we combine pixel-level and image gradient-level reprojection errors, using Huber loss ($\varepsilon$)~\cite{robust-photo, huber, huber-interpret} for image intensities and $L_1$ loss for image gradients:
\begin{equation}
\label{eq:Specific-Loss-Formulation}
\begin{array}{l}
L_{pixel}^m(u) = ||({I_1}(u) - \hat I_1^m(u)) \odot V_1^m(u)|{|_\varepsilon }\\
L_{grad}^m(u) = ||(\nabla {I_1}(u) - \nabla \hat I_1^m(u)) \odot V_1^m(u)||_1\\
L_{reproj}^m(u) = L_{pixel}^m(u) + L_{grad}^m(u).
\end{array}
\end{equation}
$V_1^m(u)$ is a visibility mask that masks out the occluded regions, where $\odot$ represents element-wise multiplication. To further reduce the influence of outlier views, we select the top-$K$ source images with the lowest reprojection error per pixel:
\begin{equation}
\label{eq:L-reproj}
L_{reproj} = \sum_u \min_{\substack{m_1, \dots, m_K \\ m_i \ne m_j \\ V_1^{m_k}(u) > 0}} \;\; \sum_{m_k} L_{reproj}^{m_k}(u).
\end{equation}

Finally, we compute the cross-view photometric consistency loss by combining the structural similarity loss $L_{SSIM}$~\cite{robust-photo, ssim-eval} calculated on visible areas of all $M$ warped source images  and the above reprojection loss: 
\begin{equation}
\label{eq:Final-photo-loss}
{L_{photo}} = {L_{reproj}} + {L_{SSIM}}.
\end{equation}
This loss serves as the objective for test-time adaptation, allowing the model to refine its depth predictions based on scene-specific geometric and photometric cues without requiring additional manual labelling. 
\subsection{Meta-Auxiliary Learning for TTA}
\noindent However, we observed that naively conducting TTA by minimizing Eq.~\ref{eq:Final-photo-loss} leads to negative knowledge transfer due to misalignment between the primary and the auxiliary task (as reported in Tab. \ref{tb:Ablation-results}). Therefore, to enhance the effectiveness of TTA, we employ a meta-auxiliary learning strategy to couple the two losses and optimize the model to benefit the primary task through self-supervised adaptation.

Specifically, we have two phases, namely meta-auxiliary training and test-time adaptation. The goal is to meta-learn the model parameters $\theta$ that can be efficiently adapted at test time through a few gradient steps with the photometric consistency loss. 

\noindent {\bf Meta Training.} We employ a MAML-style meta-training procedure~\cite{MAML,chi2025learning}, where the model is trained to improve its primary task performance—i.e., depth estimation—after being adapted using the self-supervised TTA objective. Given a batch of training samples $\{ \{ {P_{i,b}}\} _{i = 1}^{M + 1},{D_{{\rm{gt}},b}}\} _{b = 1}^B$, where $B$ denotes meta batch size, and model parameters $\theta$, the nested meta-training loop proceeds as follows.

First, in the inner loop, for the $b^{th}$ sample in the batch, we simulate test-time adaptation by updating the model using the photometric consistency loss for a few gradient steps:
\begin{equation}
\label{eq:inner-update}
{\phi _b} \leftarrow \theta  - \alpha {\nabla _\theta }{L_{photo}}(\theta ;\{ {P_{i,b}}\} _{i = 1}^{M + 1}),
\end{equation}
where $\alpha$ is the inner-loop learning rate and $M$ denotes the number of source views used for adaptation.

We want to ensure that the adapted parameter $\phi _b$ improves the primary depth estimation task. Therefore, in the outer loop, we evaluate performance on the primary task using ground-truth depth supervision and compute the meta-gradient using the adapted parameters ${\phi _b}$:
\begin{equation}
\label{eq:Outer-Update}
\theta  \leftarrow \theta  - \beta \sum\limits_{b = 1}^B {{\nabla _\theta }{L_{pri}}({\phi _b};\{ {P_{i,b}}\} _{i = 1}^N,{D_{{\rm{gt}},b}})} ,
\end{equation}
where $\beta$ is the meta-learning rate. While the primary loss is computed using the adapted parameters ${\phi _b}$ and $N-1$ source views (a subset of the $M$ views used in TTA, with $N-1 < M$), the optimization is applied to the original parameters $\theta$, encouraging them to be adaptable via self-supervised TTA.

This process enables the model to acquire a set of parameters, where small updates using the photometric loss can lead to improvements in the depth prediction task. The MAML-style optimization algorithm is summarized in Fig. \ref{fig:method_overview} and Algorithm \ref{alg:algorithm}.

\noindent {\bf Test-time Adaptation.} At inference time, the meta-trained parameters $\theta$ are adapted to a new scene by minimizing the photometric consistency loss on each test sample containing a reference image and $M$ source images for a fixed number of gradient steps:
\begin{equation}
\label{eq:tta}
\phi  \leftarrow \theta  - \alpha {\nabla _\theta }{L_{photo}}(\theta ;\{ {P_i}\} _{i = 1}^{M + 1}).
\end{equation}
The adapted model is then used to infer the depth map for the reference image. Fig. \ref{fig:tto_process} illustrates the TTA process.
\begin{algorithm}[tb]
\caption{Meta-auxiliary training}
\label{alg:algorithm}
\textbf{Require}: $\{ {P_i}\} _{i = 1}^{M + 1},{D_{{\rm{gt}}}}$: posed images with depth label\\
\textbf{Require}: $\alpha,\beta$: learning rates \\
\textbf{Output}: $\theta$: learned parameters
\begin{algorithmic}[1] 
\STATE Initialize the network with pre-trained weights $\theta$
\WHILE{not done}
\STATE Sample a training batch $\{ \{ {P_{i,b}}\} _{i = 1}^{M + 1},{D_{{\rm{gt}},b}}\} _{b = 1}^B$
\FOR{each sample}
\STATE Compute photometric consistency loss ${L_{photo}}$

\STATE Perform self-supervised adaptation : \\${\phi _b} \leftarrow \theta  - \alpha {\nabla _\theta }{L_{photo}}(\theta ;\{ {P_{i,b}}\} _{i = 1}^{M + 1})$
\ENDFOR
\STATE Evaluate the primary task using the adapted parameters and update:
\\ $\theta  \leftarrow \theta  - \beta \sum\limits_{b = 1}^B {{\nabla _\theta }{L_{pri}}({\phi _b};\{ {P_{i,b}}\} _{i = 1}^N,{D_{{\rm{gt}},b}})} $
\ENDWHILE
\STATE \textbf{return} $\theta$
\end{algorithmic}
\end{algorithm}

\section{Experiments}
\noindent In this section, we first introduce the datasets, evaluation metrics, and implementation details in \secref{wr:expr-setup}. Then, in \secref{wr:expr-dtu}, we present the main results of applying our method on DTU benchmark. \secref{wr:expr-blend} provides results of applying MVS-TTA on BlendedMVS benchmark. \secref{wr:expr-cross} presents experimental results of applying our method in a cross-dataset generalization setting. Finally, \secref{wr:expr-ablation} provides detailed ablation studies to analyze the effectiveness of each component and understand their contributions to overall performance.
\subsection{Experimental Setup}
\label{wr:expr-setup}
\noindent {\bf Datasets.} We evaluate our framework on two widely used MVS benchmarks: DTU and BlendedMVS.
DTU is a high-quality indoor dataset containing 124 scenes with ground-truth depth and calibrated camera parameters. Each scene includes 49 or 64 images. We follow the official split with 79 training and 22 test scenes.
BlendedMVS is a large-scale synthetic dataset with over 17,000 high-resolution images covering diverse outdoor scenes. Its scene diversity and unstructured camera trajectories make it well-suited for evaluating generalization under challenging domain shifts.

\noindent {\bf Evaluation Metrics.} 
We evaluate depth estimation quality using two commonly adopted metrics: absolute relative depth error (rel, $\downarrow$ lower is better) and inlier percentage ($\tau$, $\uparrow$ higher is better), following the evaluation protocol in MVSAnywhere~\cite{mvsanywhere}. 
We report results at $t=1.03$ and include a more relaxed threshold at $t=1.10$ to better capture near-correct predictions. 

\noindent {\bf Implementation Details.} For fair and faithful comparison, we integrate our method with the prior works whose official model checkpoints are released. During meta-auxiliary training, the meta-learning rate $\beta$ and the adaptation learning rate $\alpha$ are set to $10^{-4}$. For all backbones, meta-training is conducted for 200 iterations. All experiments are conducted on a single NVIDIA V100 GPU. The inference time of GeoMVSNet is approximately 0.45\,s per test sample, and applying 2 steps of test-time adaptation adds $\sim$1.0\,s of additional computation. To support gradient-based TTA, for models using classification-based depth prediction (e.g., softmax over discrete depth bins)~\cite{mvsformer++, geomvsnet, transmvsnet}, we replace the classification computation with expectation-based depth formulation, i.e., computing the expected depth over discrete hypotheses~\cite{cascade}, enabling adaptation while preserving the original model architecture.
\begin{figure}[t]
\scriptsize  
\renewcommand{\arraystretch}{0.1} 
\setlength{\tabcolsep}{1pt}     
\centering
\begin{tabular}{cccc}
        & \textbf{MVSFormer++} & \textbf{GeoMVSNet} & \textbf{TransMVSNet} \\
        \rotatebox{90}{\parbox{1.9cm}{\centering \textbf{Reference}\\\textbf{Image}}} &
        \includegraphics[height=1.9cm,valign=b]{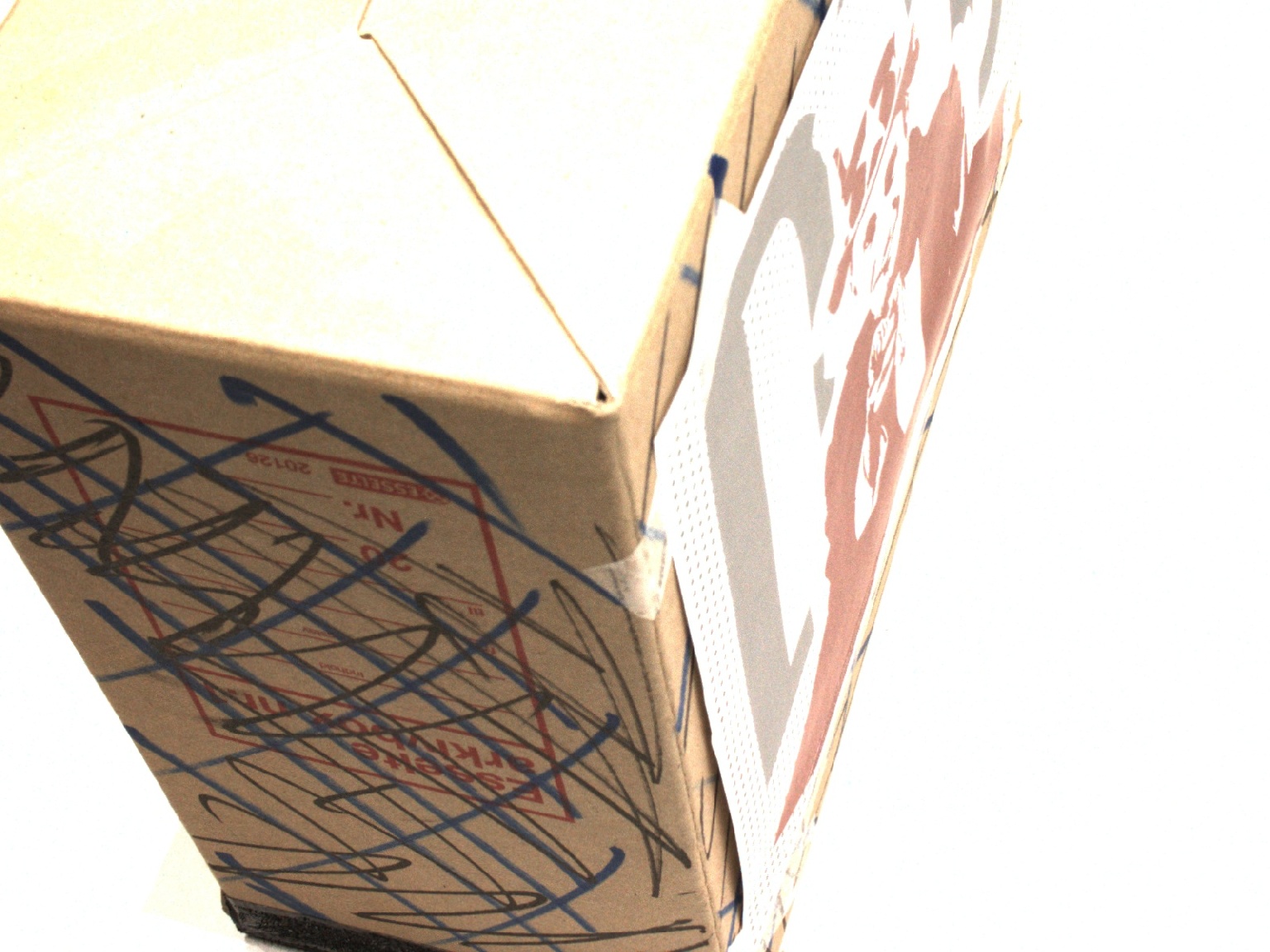} &
        \includegraphics[height=1.9cm,valign=b]{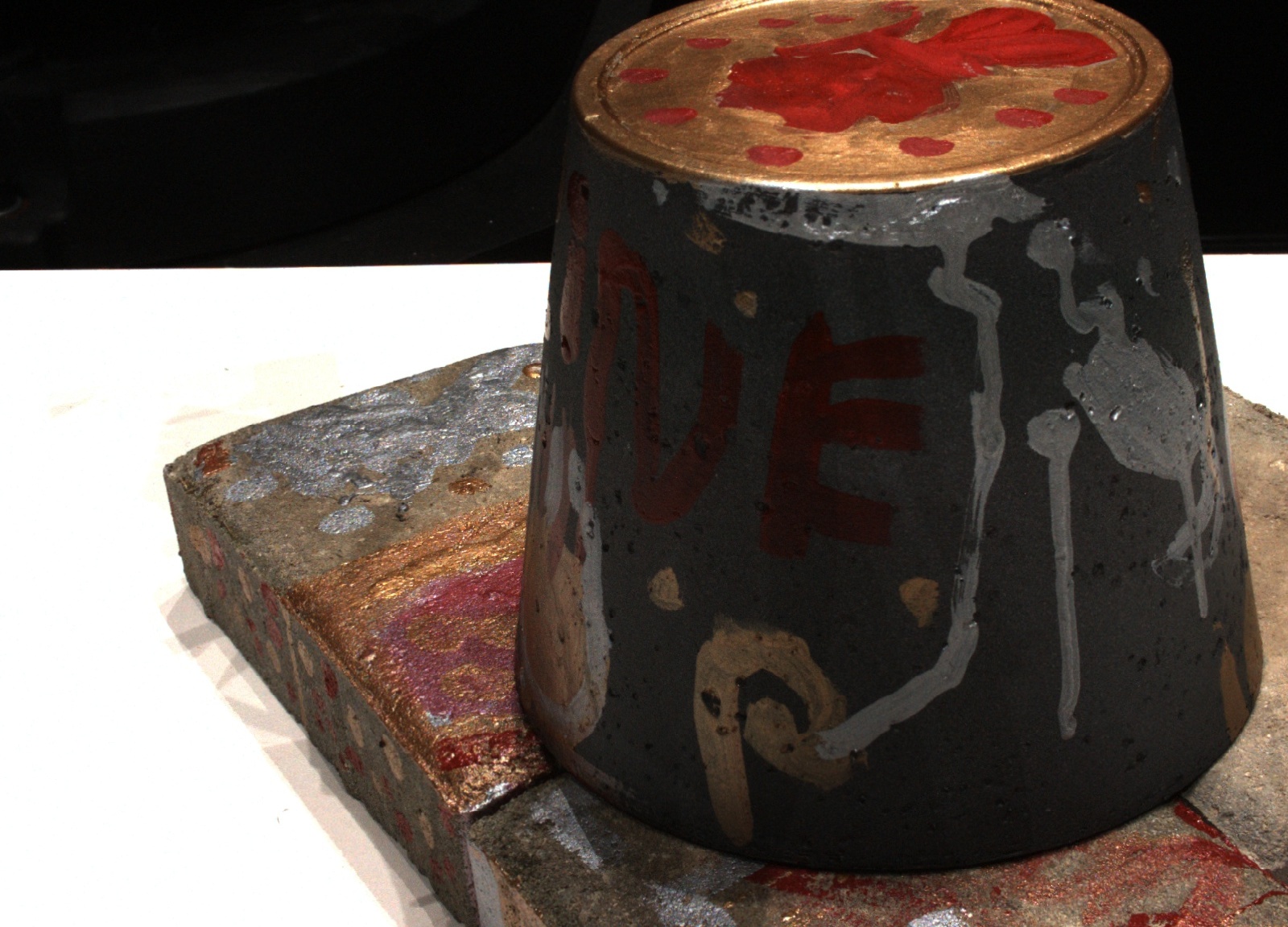} &
        \includegraphics[height=1.9cm,valign=b]{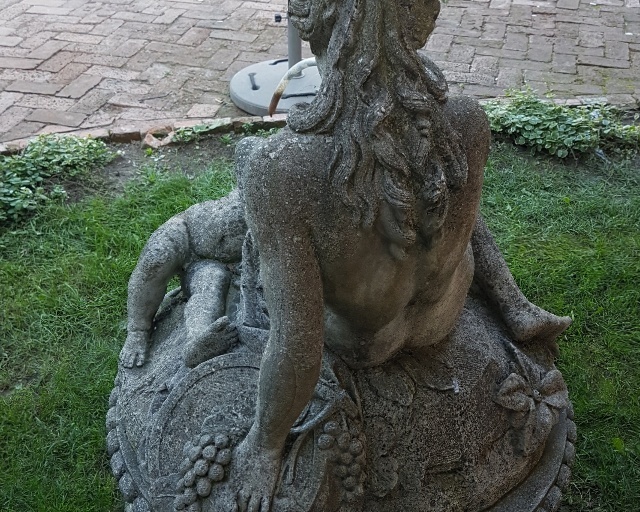} \\
        \rotatebox{90}{\parbox{1.9cm}{\centering \textbf{Ground Truth}\\\textbf{Depth}}} &
        \includegraphics[height=1.9cm,valign=b]{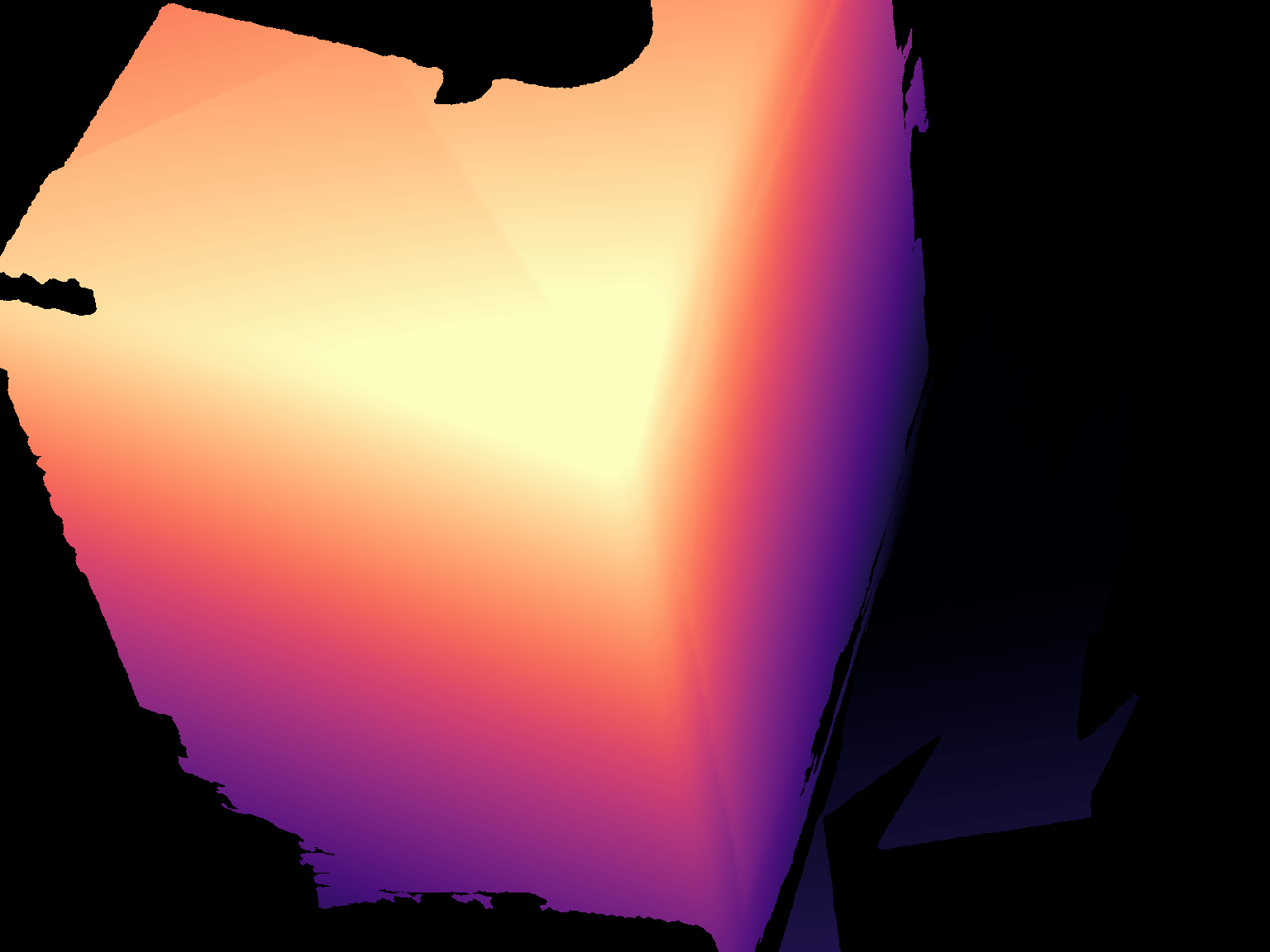} &
        \includegraphics[height=1.9cm,valign=b]{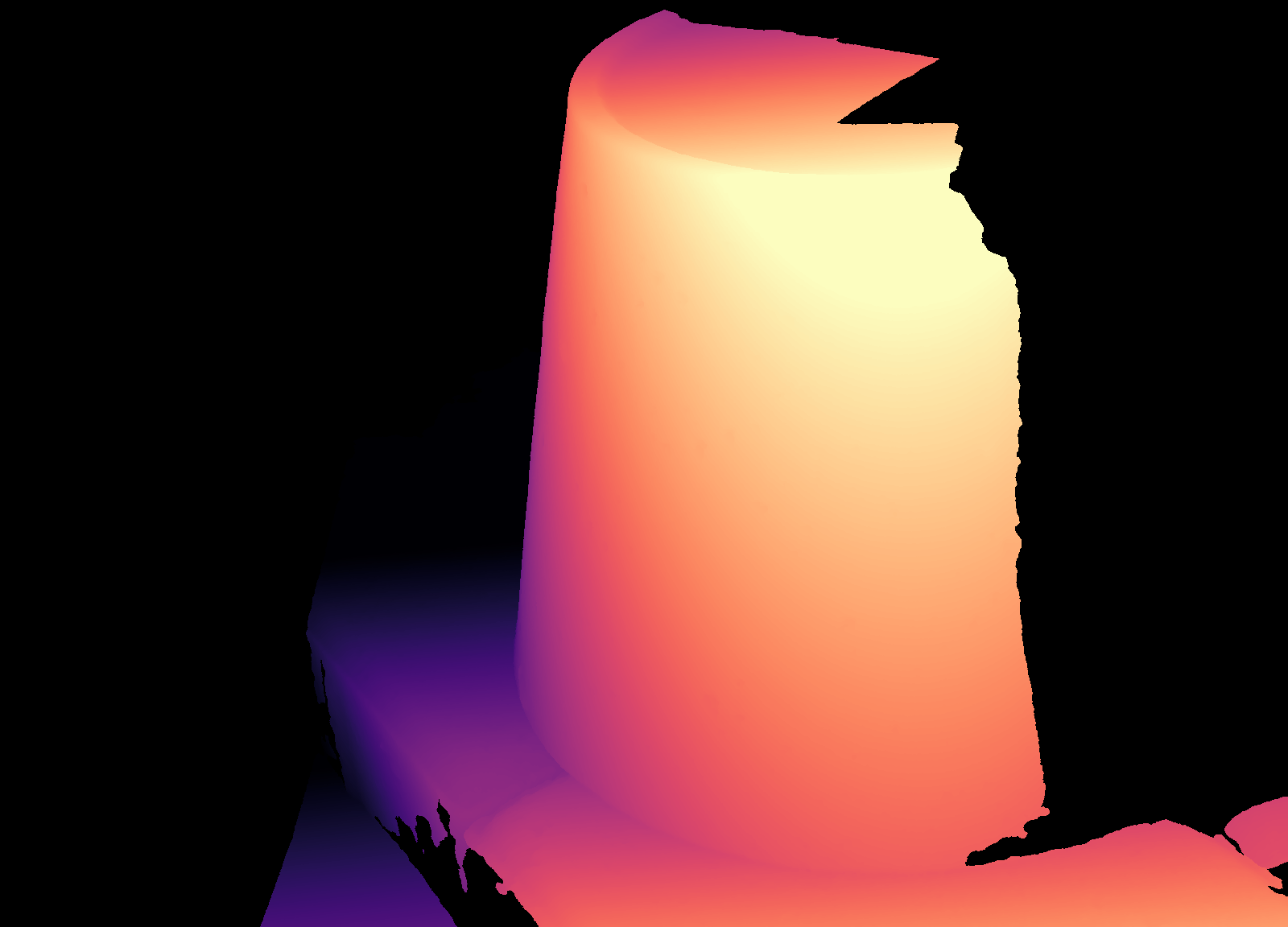} &
        \includegraphics[height=1.9cm,valign=b]{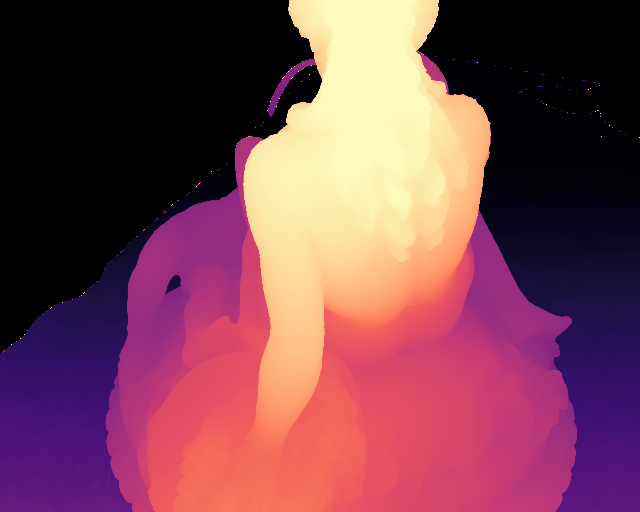} \\
        \rotatebox{90}{\parbox{1.9cm}{\centering \textbf{Baseline}}} &
        \includegraphics[height=1.9cm,valign=b]{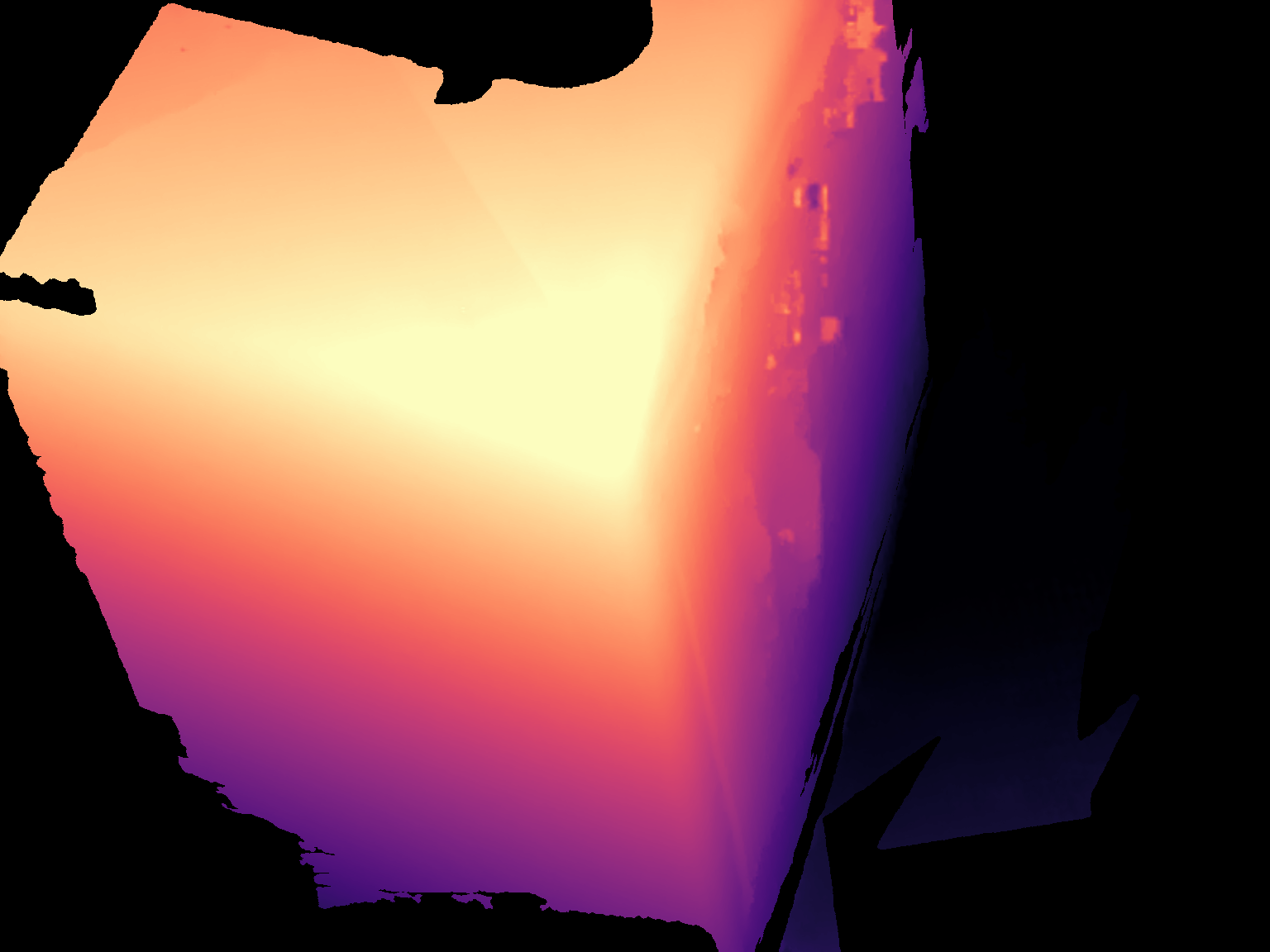} &
        \includegraphics[height=1.9cm,valign=b]{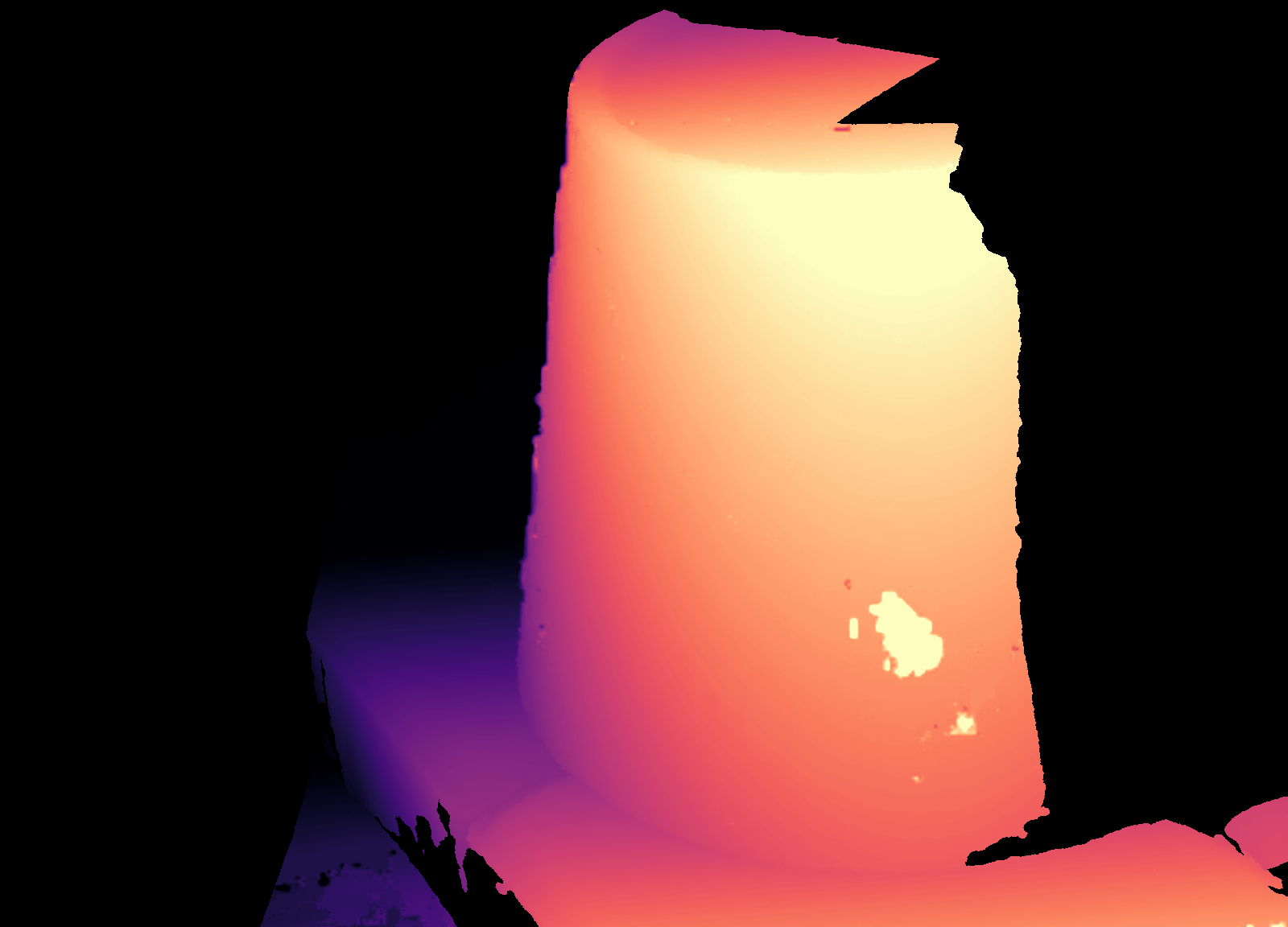} &
        \includegraphics[height=1.9cm,valign=b]{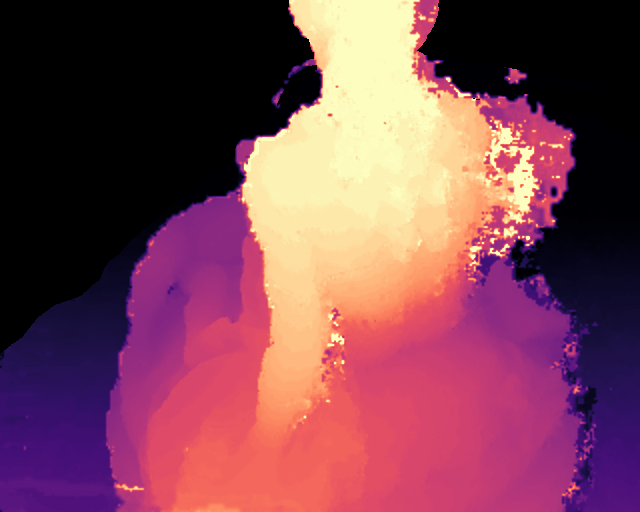} \\
        \rotatebox{90}{\parbox{1.9cm}{\centering \textbf{Baseline + MVS-TTA}}} &
        \includegraphics[height=1.9cm,valign=b]{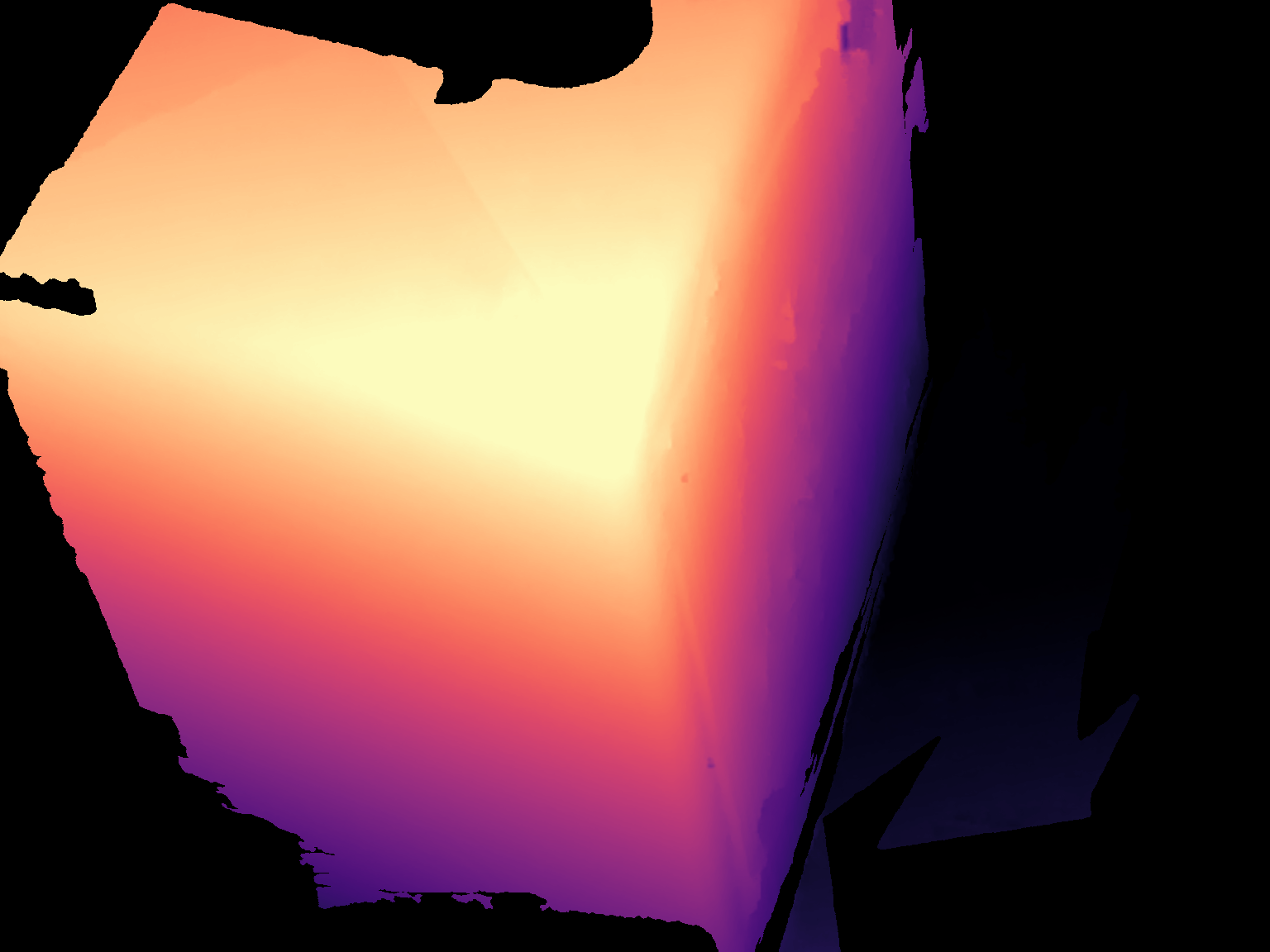} &
        \includegraphics[height=1.9cm,valign=b]{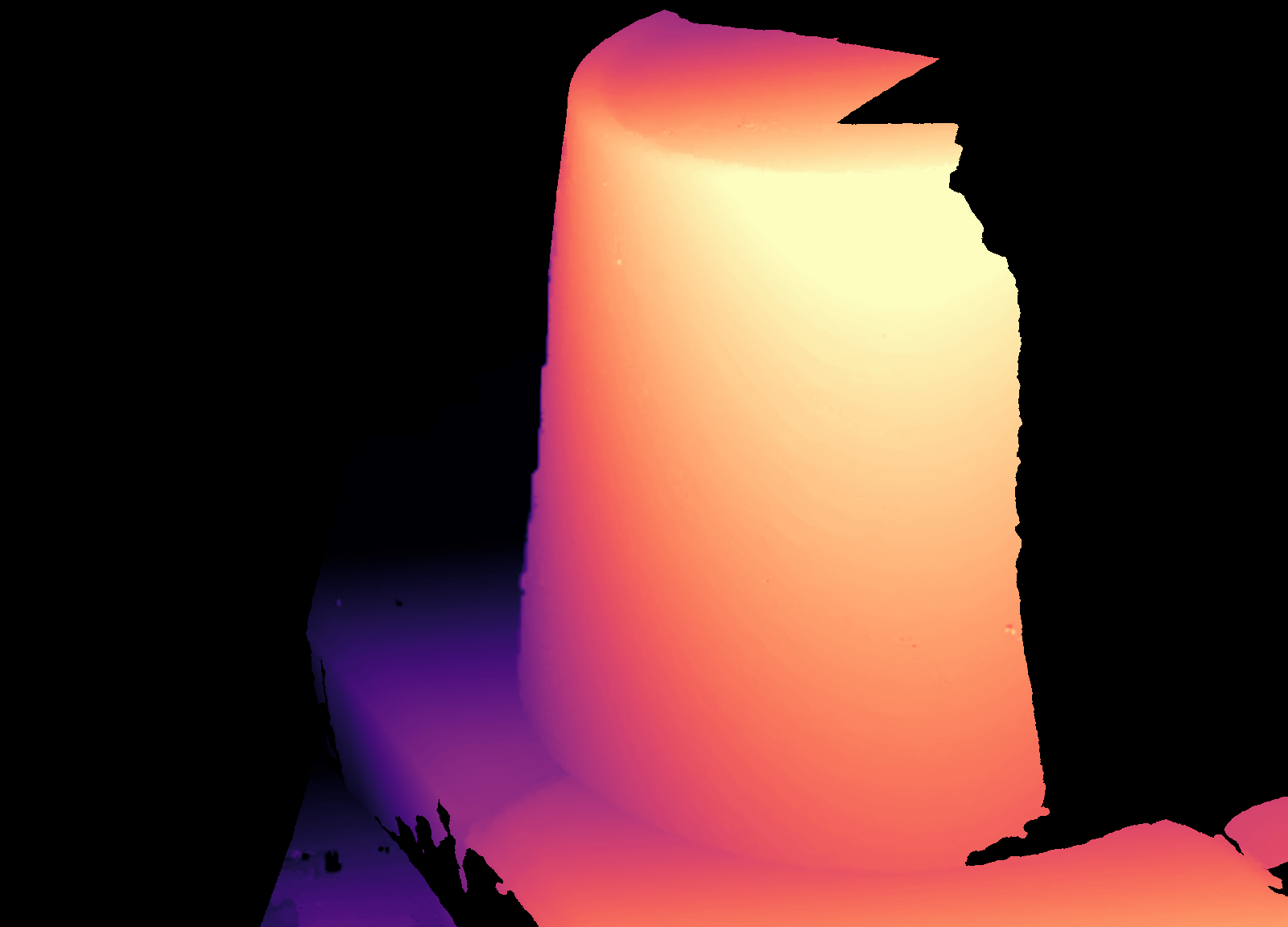} &
        \includegraphics[height=1.9cm,valign=b]{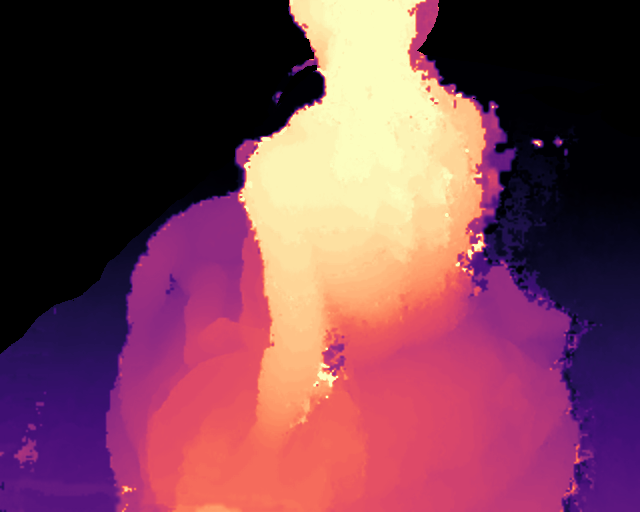} \\
\end{tabular}
\caption{Qualitative comparison between different baselines and our MVS-TTA framework. Each column corresponds to a different MVS model and dataset combination: the first two columns show results on the DTU dataset, and the last column is from BlendedMVS. From top to bottom: reference image, ground-truth depth map, depth prediction from the baseline model, and depth prediction after applying MVS-TTA. Our method improves the accuracy of depth prediction across different models and datasets.}
\label{fig:qualitative}
\end{figure}
\subsection{Benchmarking on DTU dataset}
\label{wr:expr-dtu}
\noindent  We evaluate the effectiveness of our method on the test split of the DTU dataset. All models are trained on the DTU training set using 5 input views and 192 depth hypotheses.

As shown in Tab. \ref{tb:DTU-results}, our MVS-TTA framework consistently improves the performance of multiple learning-based MVS baselines. For instance, applying MVS-TTA to MVSFormer++ reduces the rel error from 0.91 to 0.84, while increasing $\tau$(1.03) from 95.5\% to 96.1\%. Similar improvements are observed for GeoMVSNet, TransMVSNet, and CasMVSNet, demonstrating that our framework is compatible with diverse MVS architectures. Results of reference methods are included for comparison.

Fig. \ref{fig:qualitative-textureless} focuses on three textureless scenes from the DTU benchmark, while fig. \ref{fig:qualitative} evaluates MVS-TTA across multiple MVS architectures. These results indicate that MVS-TTA enables existing models to adapt more effectively at inference time, achieving better accuracy without requiring any additional supervision even in the absence of distinctive visual cues.
\begin{table}[t]
  \centering
  \caption{Quantitative comparison on the DTU benchmark}
  \label{tb:DTU-results}
  \begin{tabularx}{\linewidth}{l *{3}{>{\centering\arraybackslash}X}}
\toprule
Method           & rel$\downarrow$             & $\tau$(1.03)$\uparrow$           & $\tau$(1.1)$\uparrow$                      \\
\midrule
Patchmatchnet~\cite{patchmatchnet}    & 2.1                           & 82.8                                               & -                                                 \\
Vis-MVSNet~\cite{vis-mvsnet}       & 1.8                           & 87.4                                               & -                                                 \\
MVSNet           & 1.8                           & 86.0                                               & -  \\  
\midrule
MVSFormer++      & 0.91                         & 95.5                                               & 97.4                                              \\
\rowcolor{gray!20} + MVS-TTA (ours) & 0.84                         & {\footnotesize 96.1 {\scriptsize (+0.6)}}                                              & 97.7                                              \\
GeoMVSNet        & 2.08                          & 88.3                                               & 92.9                                              \\
\rowcolor{gray!20} + MVS-TTA (ours)   & 1.84                          & {\footnotesize 88.6 {\scriptsize (+0.3)}}                                              & 93.7                                              \\
TransMVSNet      & 4.01                          & 78.8                                               & 86.1                                              \\
\rowcolor{gray!20} + MVS-TTA (ours) & 3.87                          & {\footnotesize 80.5 {\scriptsize (+1.7)}}                                              & 87.1                                              \\
CasMVSNet        & 1.64                          & 88.9                                               & 95.1                                              \\
\rowcolor{gray!20} + MVS-TTA (ours)   & 1.50                          & {\footnotesize 89.3 {\scriptsize (+0.4)}}                                              & 95.6                                              \\
\rowcolor{gray!20} + L2A(Tonioni et al.~\cite{tonioni2019learning}) & 1.68                          & {\footnotesize 89.1 {\scriptsize (+0.2)}}                                              & 94.9                                              \\
\bottomrule
  \end{tabularx}
\end{table}
\subsection{Evaluation on BlendedMVS dataset}
\label{wr:expr-blend}
\noindent We further evaluate our method on the validation split of the BlendedMVS dataset to assess its generalization ability under greater appearance and structural diversity. As shown in Tab. \ref{tb:Blended-results}, our test-time adaptation framework consistently improves the performance of various MVS backbones across all evaluation metrics.

Specifically, applying MVS-TTA to MVSFormer++ reduces the relative depth error from 1.46 to 1.40, while maintaining comparable inlier accuracy. For TransMVSNet, the rel error drops from 1.70 to 1.56, and $\tau$(1.03) increases from 91.0\% to 91.3\%. CasMVSNet benefits similarly, with $\tau$(1.03) improving from 91.5\% to 92.2\% through TTA. The gain on MVSFormer++ is modest since its large-capacity transformer backbone and extensive pretraining already provide strong cross-scene generalization.

Fig. \ref{fig:qualitative} shows a qualitative comparison on the validation split of the BlendedMVS benchmark; these results confirm that MVS-TTA can generalize beyond indoor tabletop scenes.
\begin{table}[t]
  \centering
  \caption{Quantitative comparison on the BlendedMVS benchmark\tablefootnote{GeoMVSNet is excluded because no pretrained models or BlendedMVS results are publicly available, which makes reliable reproduction infeasible.}}
  \label{tb:Blended-results}
  \begin{tabularx}{\linewidth}{l *{3}{>{\centering\arraybackslash}X}}
\toprule
Method           & rel$\downarrow$             & $\tau$(1.03)$\uparrow$           & $\tau$(1.1)$\uparrow$                      \\
\midrule
MVSFormer++      & 1.46                          & 91.2                                               & 97.1                                              \\
\rowcolor{gray!20} + MVS-TTA (ours) & 1.40                          & {\footnotesize 91.2 {\scriptsize (+0.0)}}                                               & 97.0                                              \\
TransMVSNet      & 1.70                          & 91.0                                               & 96.2                                              \\
\rowcolor{gray!20} + MVS-TTA (ours) & 1.56                          & {\footnotesize 91.3 {\scriptsize (+0.3)}}                                               & 96.6                                              \\
CasMVSNet        & 1.81                          & 91.5                                               & 96.4                                              \\
\rowcolor{gray!20} + MVS-TTA (ours)   & 1.65                          & {\footnotesize 92.2 {\scriptsize (+0.7)}}                                               & 96.7                                              \\
\bottomrule
  \end{tabularx}
\end{table}
\subsection{Cross-dataset Generalization}
\label{wr:expr-cross}
\begin{figure}[t]
\scriptsize  
\setlength{\tabcolsep}{1pt}     
\renewcommand{\arraystretch}{0.1} 
\centering
\begin{tabular}{cccc}
        & \parbox{2.2cm}{\centering \textbf{Reference}\\\textbf{Image}} & {\parbox{2.2 cm}{\centering \textbf{Baseline}}} & {\parbox{2.2 cm}{\centering \textbf{Baseline+}\\\textbf{MVS-TTA}}} \\
        \rotatebox{90}{\parbox{1.9cm}{\centering \textbf{TransMVSNet}}} &
        \includegraphics[height=1.9cm,valign=b]{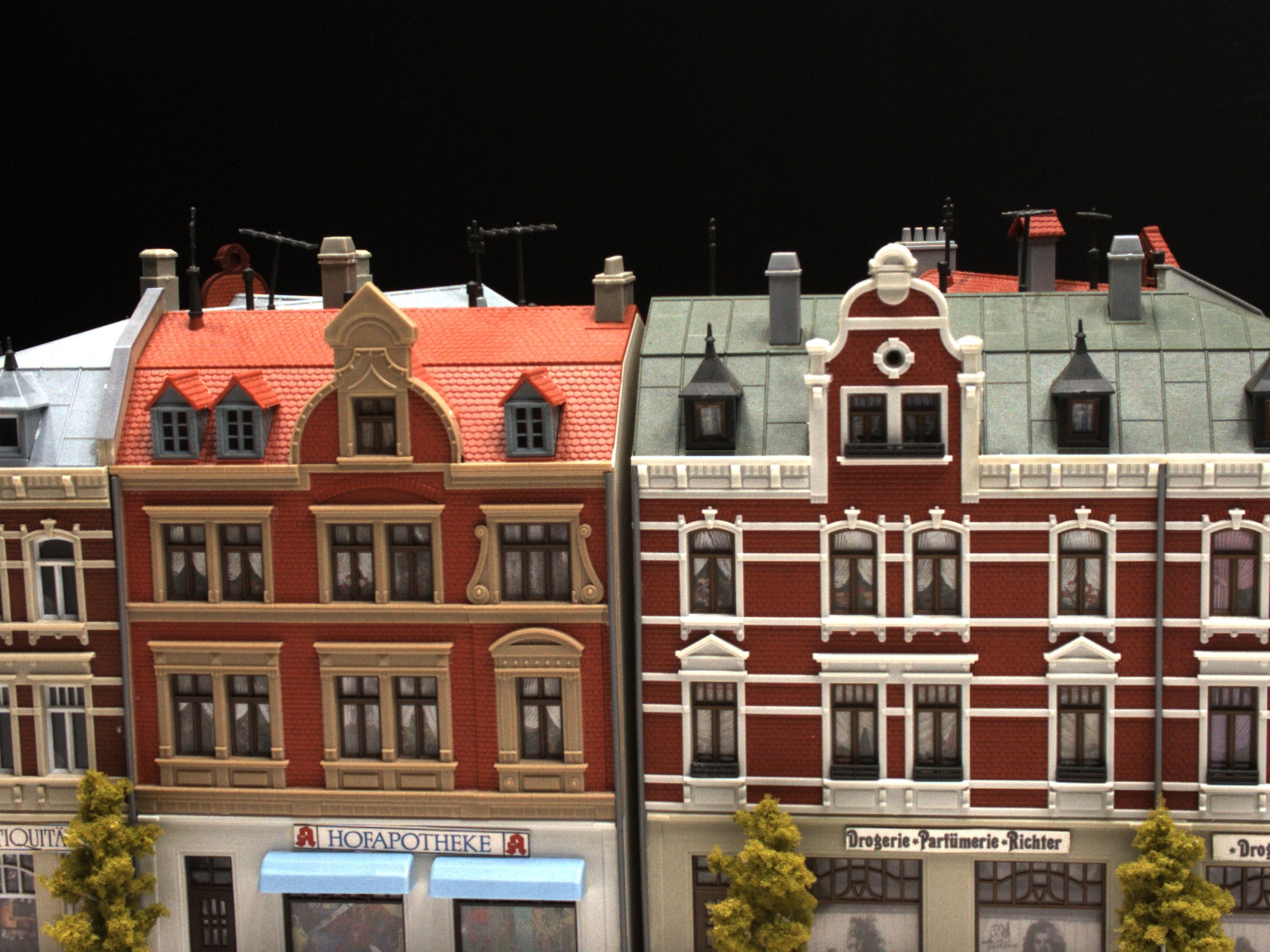} &
        \includegraphics[height=1.9cm,valign=b]{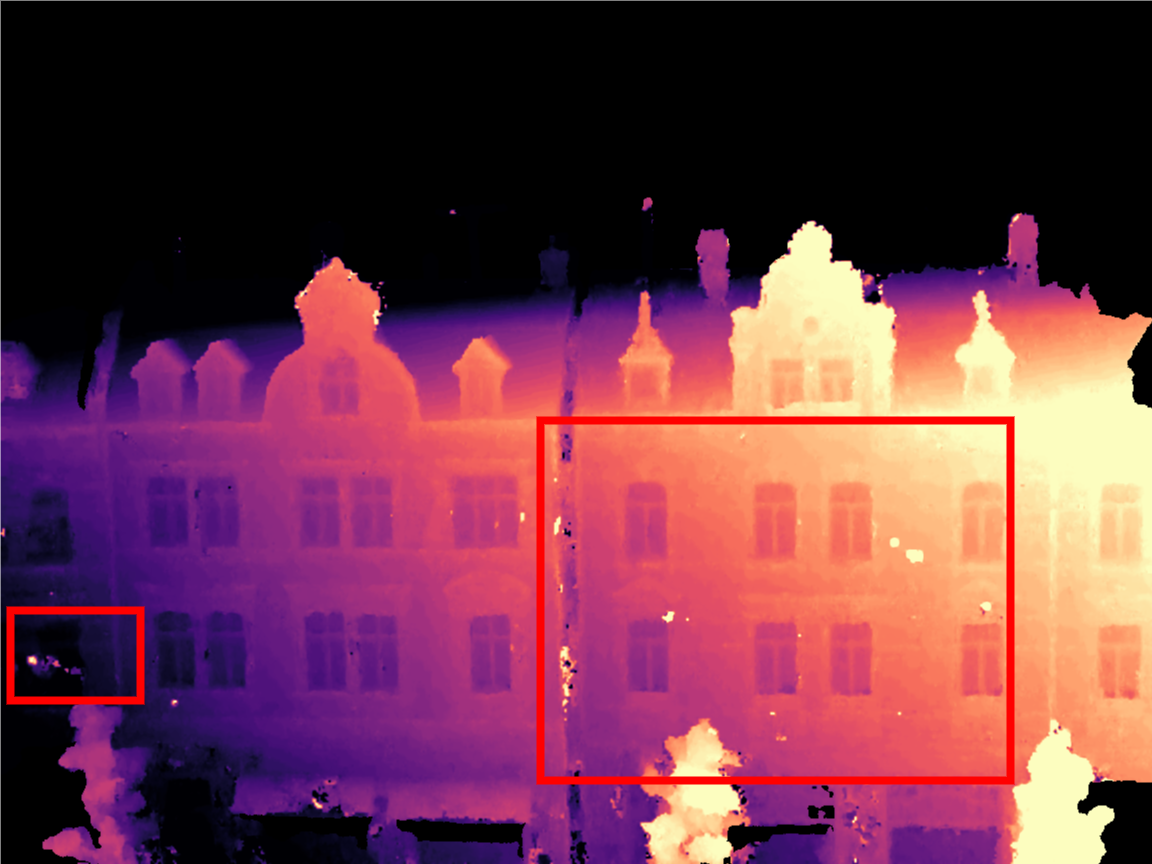} &
        \includegraphics[height=1.9cm,valign=b]{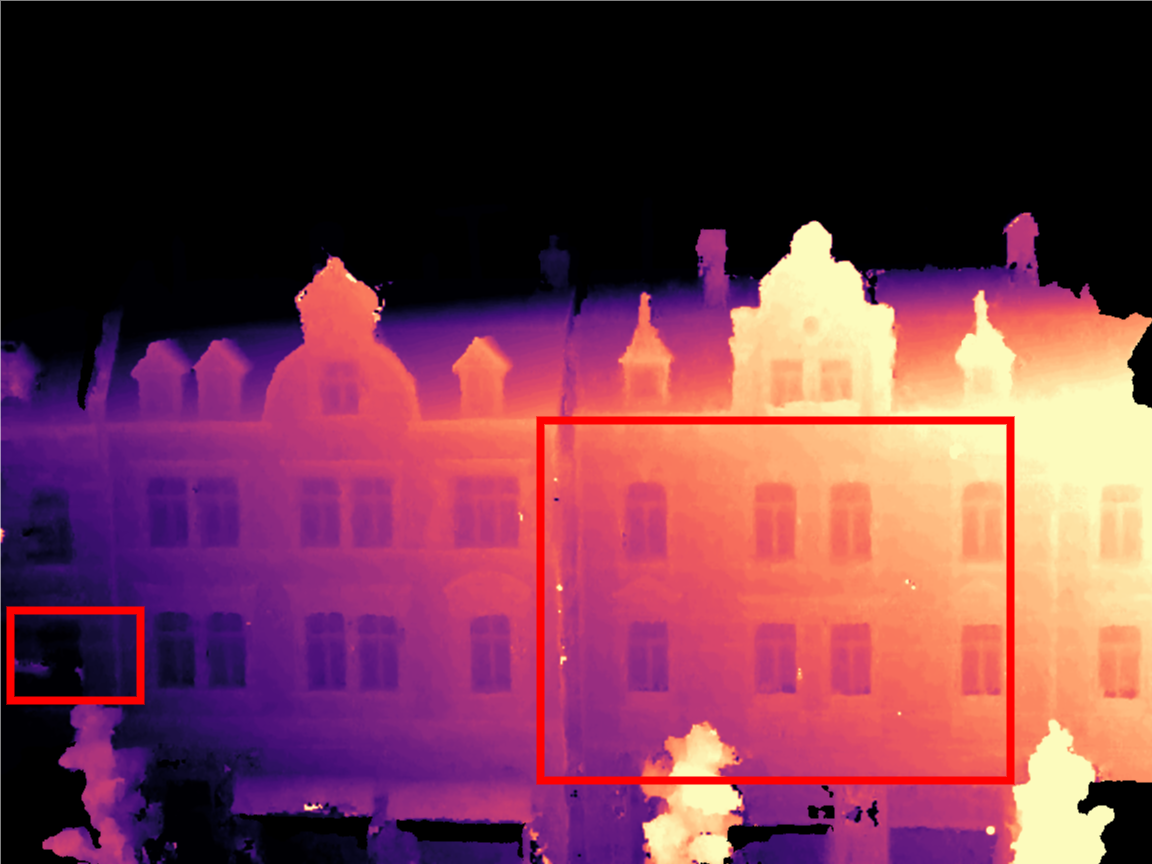} \\
        \rotatebox{90}{\parbox{1.9cm}{\centering \textbf{CasMVSNet}}} &
        \includegraphics[height=1.9cm,valign=b]{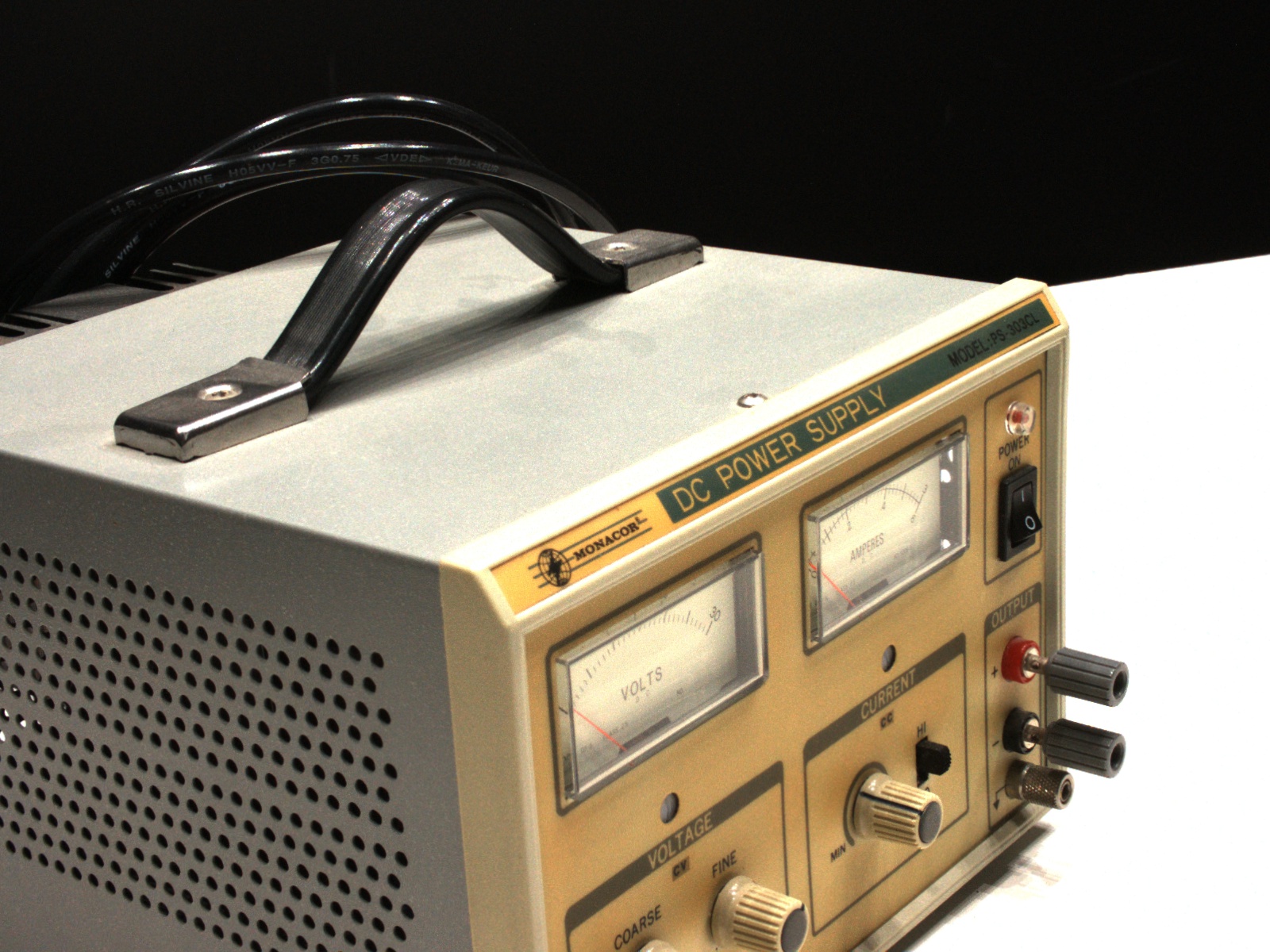} &
        \includegraphics[height=1.9cm,valign=b]{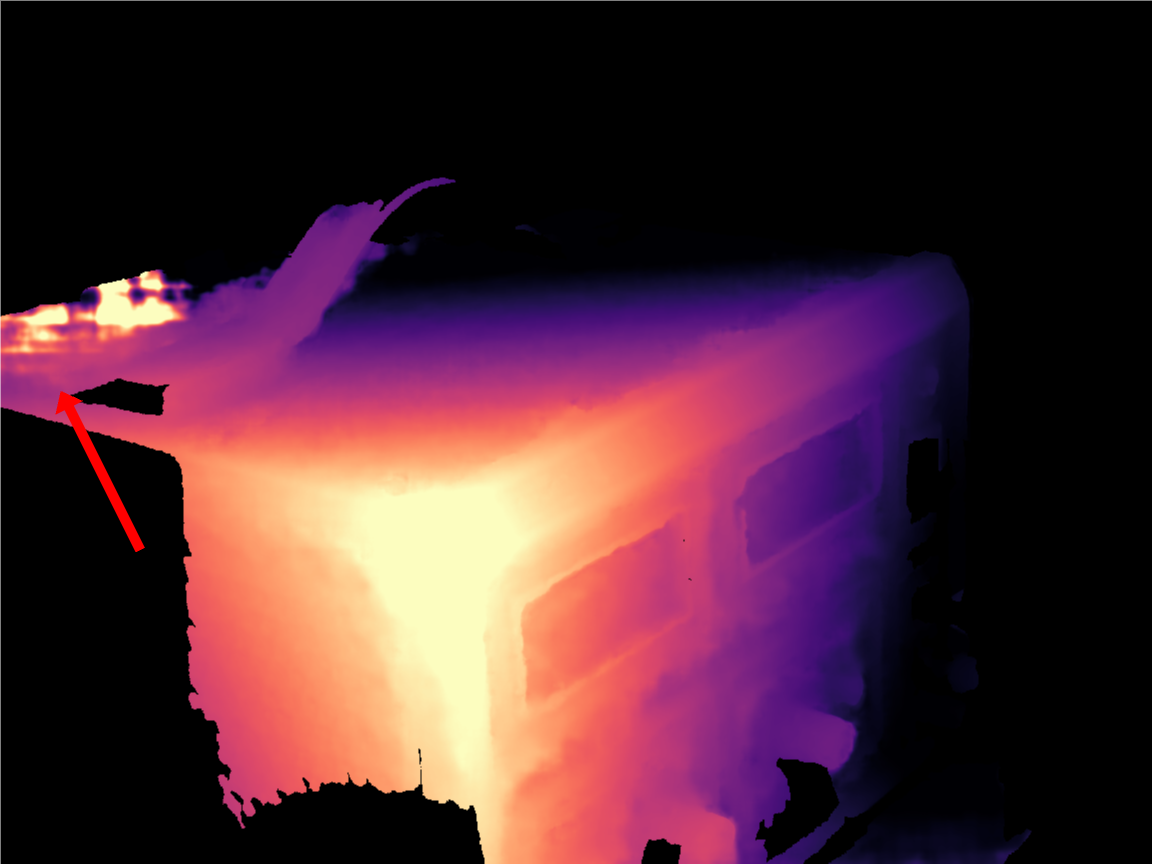} &
        \includegraphics[height=1.9cm,valign=b]{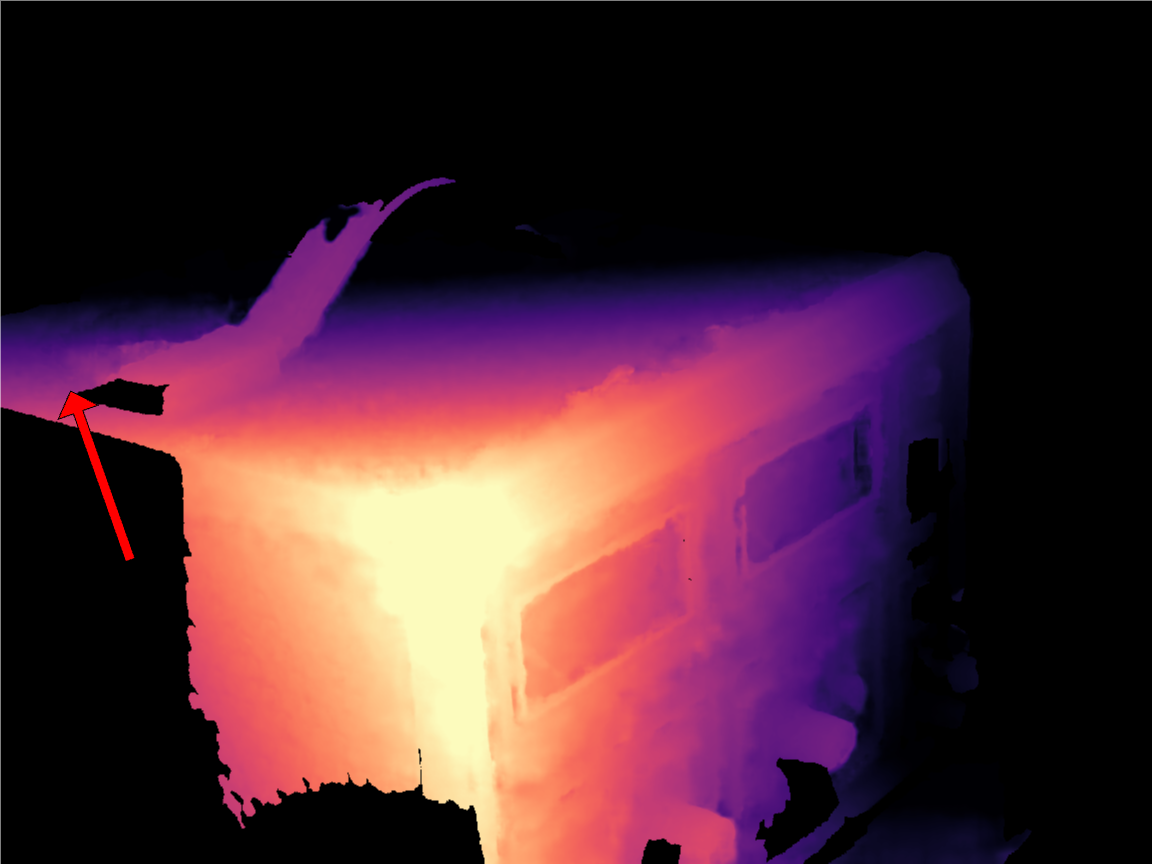} \\
\end{tabular}
\caption{Qualitative comparison of cross-dataset generalization. The first row shows results using TransMVSNet as backbone, and the second row presents results using CasMVSNet as backbone. From left to right: reference image, depth prediction using baseline model without any adaptation, and depth prediction with MVS-TTA applied. In the first row, the scattered erroneous predictions in front of the building façade are reduced, and in the second row, the incorrect depth around the upper-left region of the instrument is corrected.}
\label{fig:cross-qualitative}
\end{figure}

To assess the generalization ability of our framework under significant distribution shifts, we conduct cross-dataset experiments by training the models on the BlendedMVS dataset and directly evaluating them on the DTU test set, without any fine-tuning. Tab. \ref{tb:Cross-Dataset-results} reports the absolute relative depth error and inlier percentages for CasMVSNet and TransMVSNet, while Fig. \ref{fig:cross-qualitative} presents corresponding qualitative evaluation. Our MVS-TTA framework consistently improves generalization performance across both backbones. For CasMVSNet, the absolute relative depth error decreases from 4.31 to 4.10, while $\tau$(1.03) and $\tau$(1.1) increase to 82.5\% and 86.2\%, respectively. TransMVSNet sees a more notable gain, with the rel error reduced from 4.81 to 4.35, and $\tau$(1.03) improving from 79.1\% to 79.8\%, indicating more robust depth predictions on unseen domains. These results demonstrate that MVS-TTA not only improves performance in standard within-dataset settings but also enhances the transferability of MVS models across datasets with different scene characteristics and visual distributions, without relying on ground-truth depth.
\begin{table}[t]
  \centering
  \caption{Cross-dataset generalization performance}
  \label{tb:Cross-Dataset-results}
  \begin{tabularx}{\linewidth}{l *{3}{>{\centering\arraybackslash}X}}
\toprule
Method           & rel$\downarrow$             & $\tau$ (1.03)$\uparrow$           & $\tau$(1.1)$\uparrow$                      \\
\midrule
CasMVSNet        & 4.31                          & 82.2                                               & 86.0                                              \\
\rowcolor{gray!20} + MVS-TTA (ours)   & 4.10                          & 82.5                                               & 86.2                                              \\
TransMVSNet        & 4.81                          & 79.1                                               & 83.1                                              \\
\rowcolor{gray!20} + MVS-TTA (ours)   & 4.35                          & 79.8                                               & 84.4                                              \\
\bottomrule
  \end{tabularx}
\end{table}
\subsection{Ablation Study}
\label{wr:expr-ablation}
\begin{figure*}[t]
\scriptsize  
\renewcommand{\arraystretch}{0.1} 
\setlength{\tabcolsep}{1pt} 
\centering
\begin{tabular}{cccccc}
        \parbox{2.2cm}{\centering \textbf{Reference}\\\textbf{Image}} & {\parbox{2.2 cm}{\centering \textbf{Ground Truth}\\\textbf{Depth}}} & {\parbox{2.2 cm}{\centering \textbf{CasMVSNet \\\textbf{(Baseline)}}}} & {\parbox{2.2 cm}{\centering \textbf{CasMVSNet+}\\\textbf{TTA}}} & {\parbox{2.2 cm}{\centering \textbf{CasMVSNet+}\\\textbf{Meta-auxiliary Training}}} & {\parbox{2.2 cm}{\centering \textbf{CasMVSNet+}\\\textbf{MVS-TTA}}} \\
        \includegraphics[height=2.0cm,valign=b]{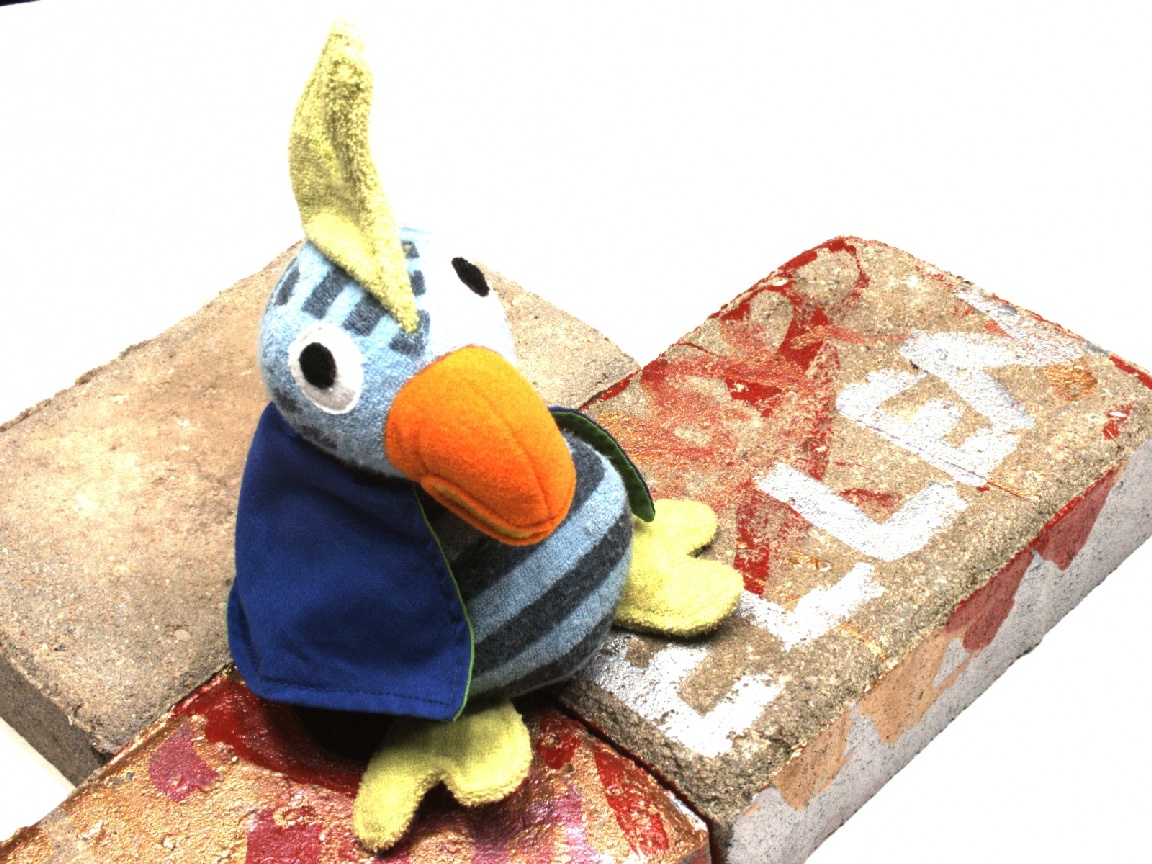} &
        \includegraphics[height=2.0cm,valign=b]{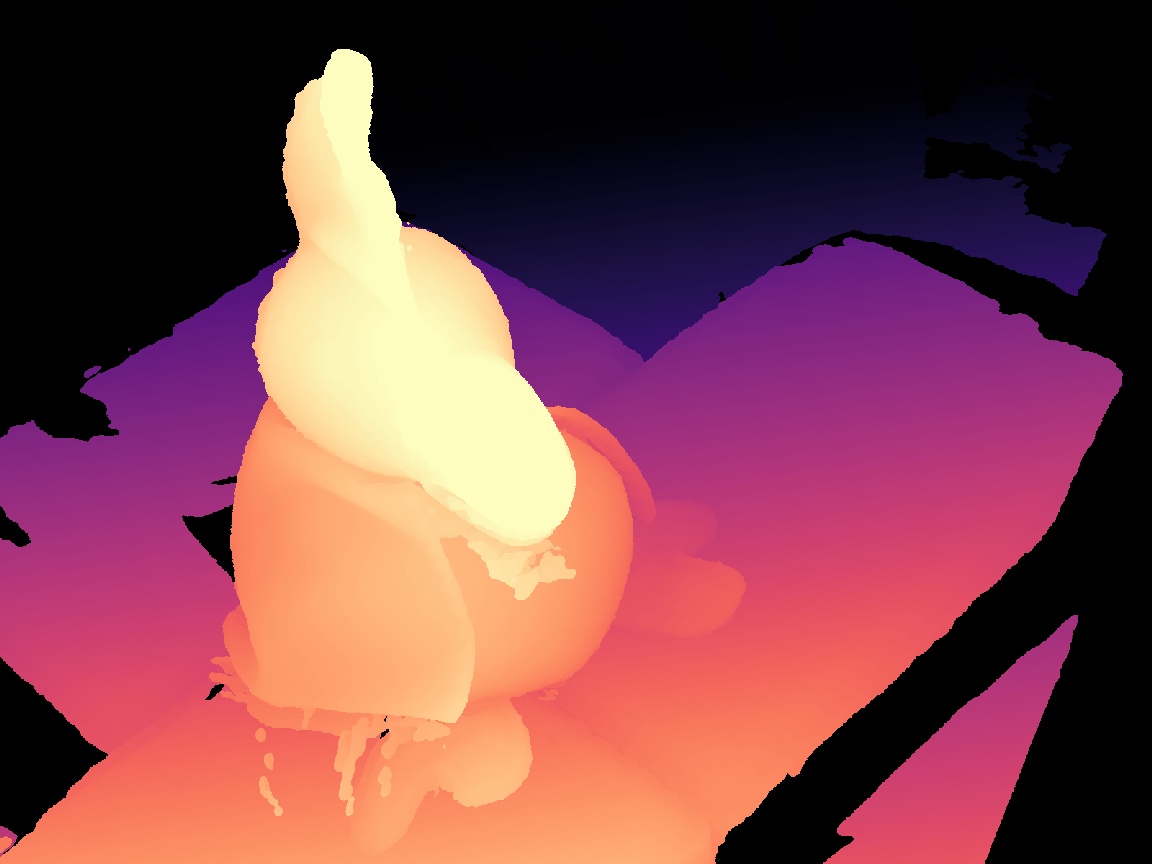} &
        \includegraphics[height=2.0cm,valign=b]{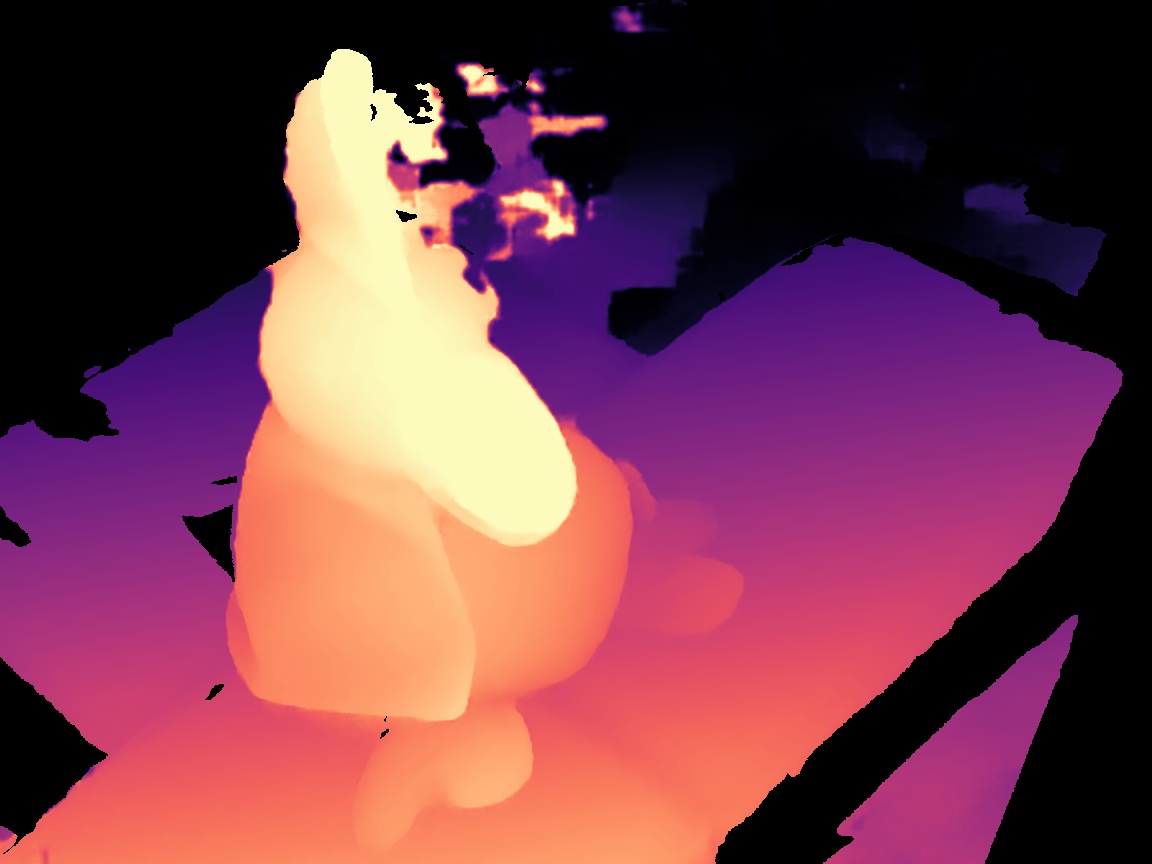} &
        \includegraphics[height=2.0cm,valign=b]{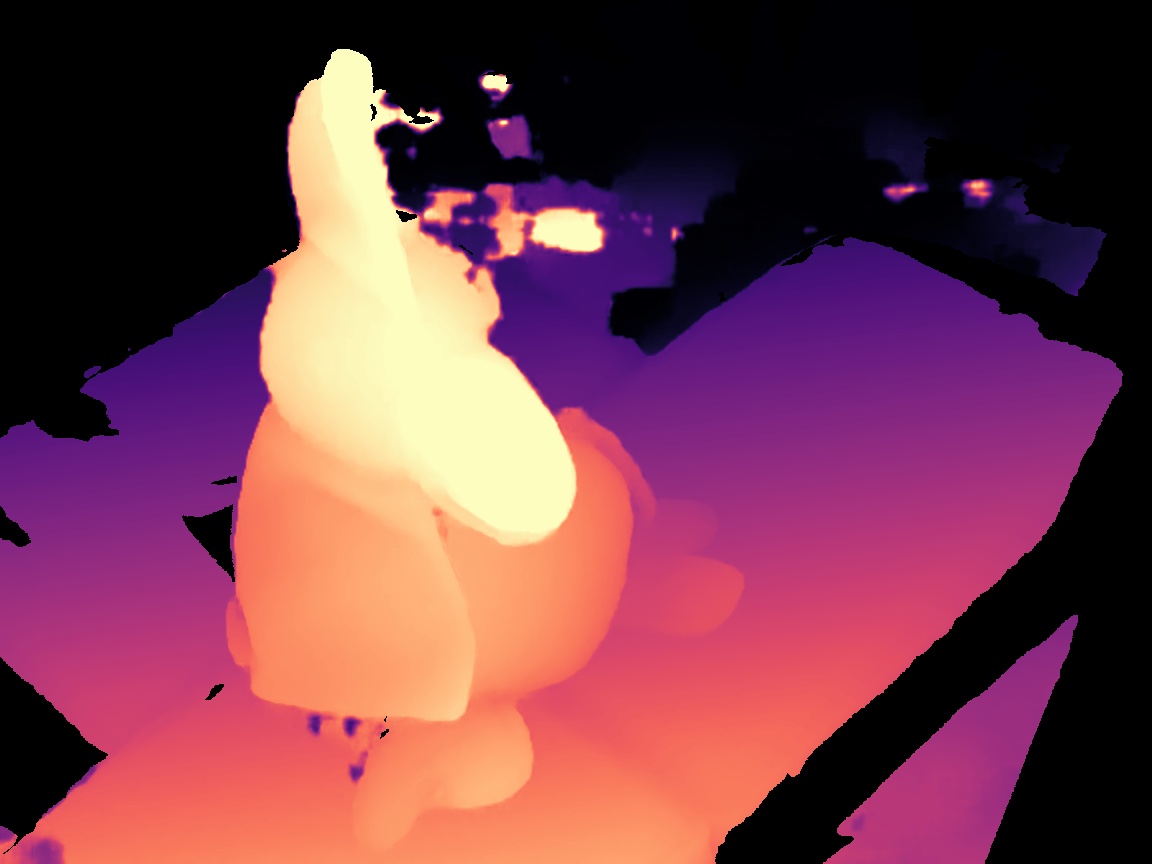} &
        \includegraphics[height=2.0cm,valign=b]{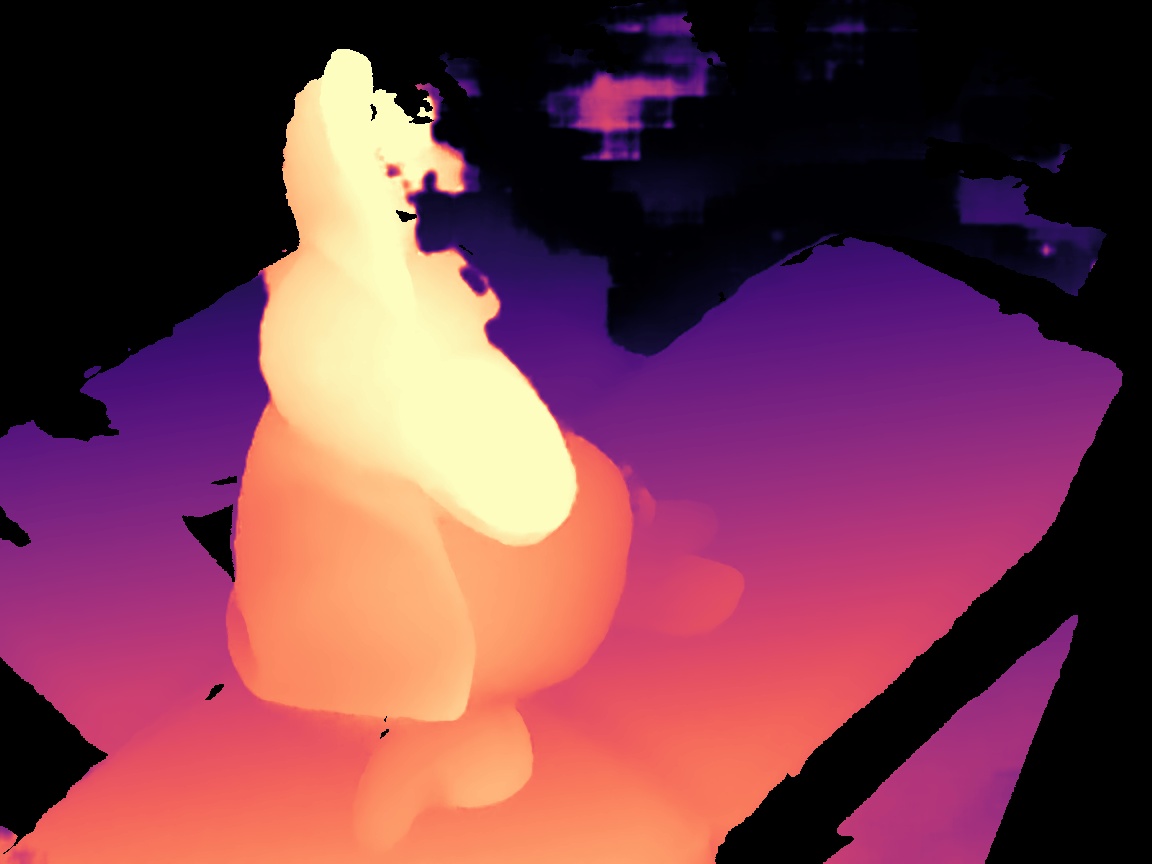} &
        \includegraphics[height=2.0cm,valign=b]{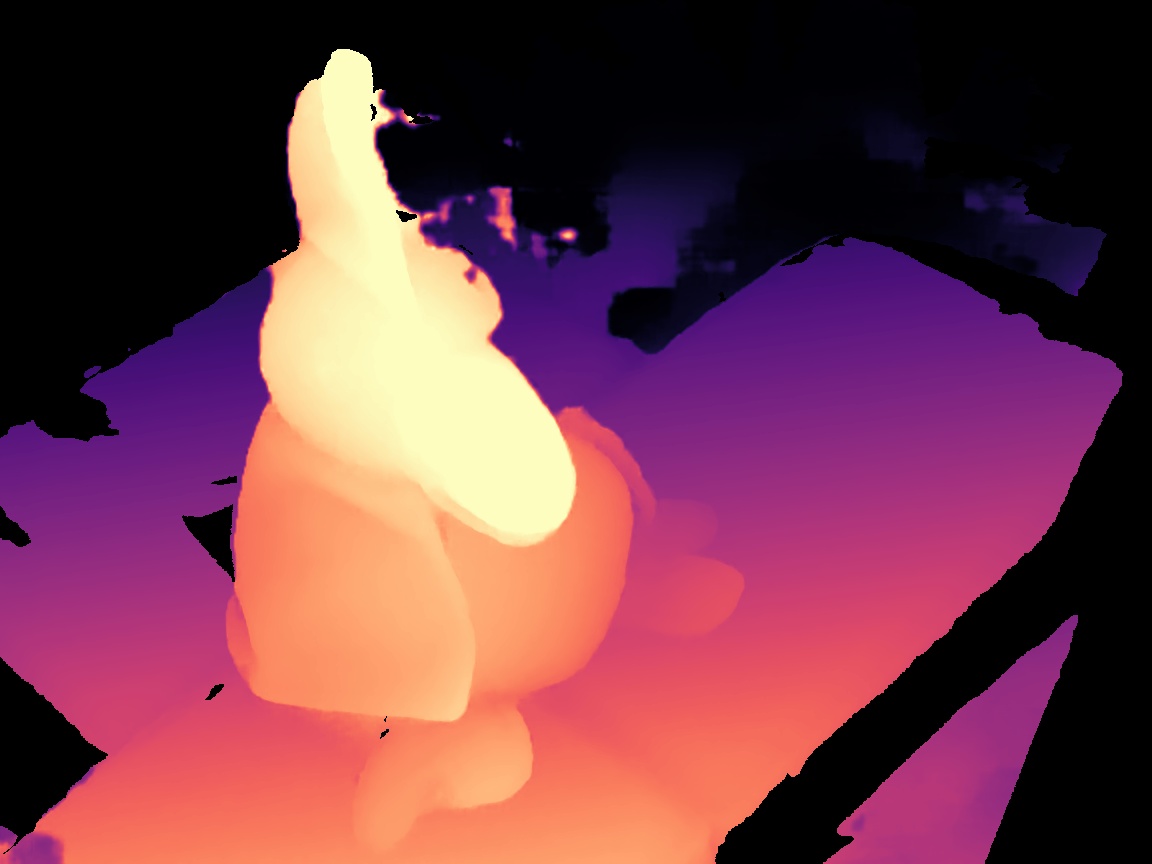} \\
        [0.5em]  
        \includegraphics[height=2.0cm,valign=b]{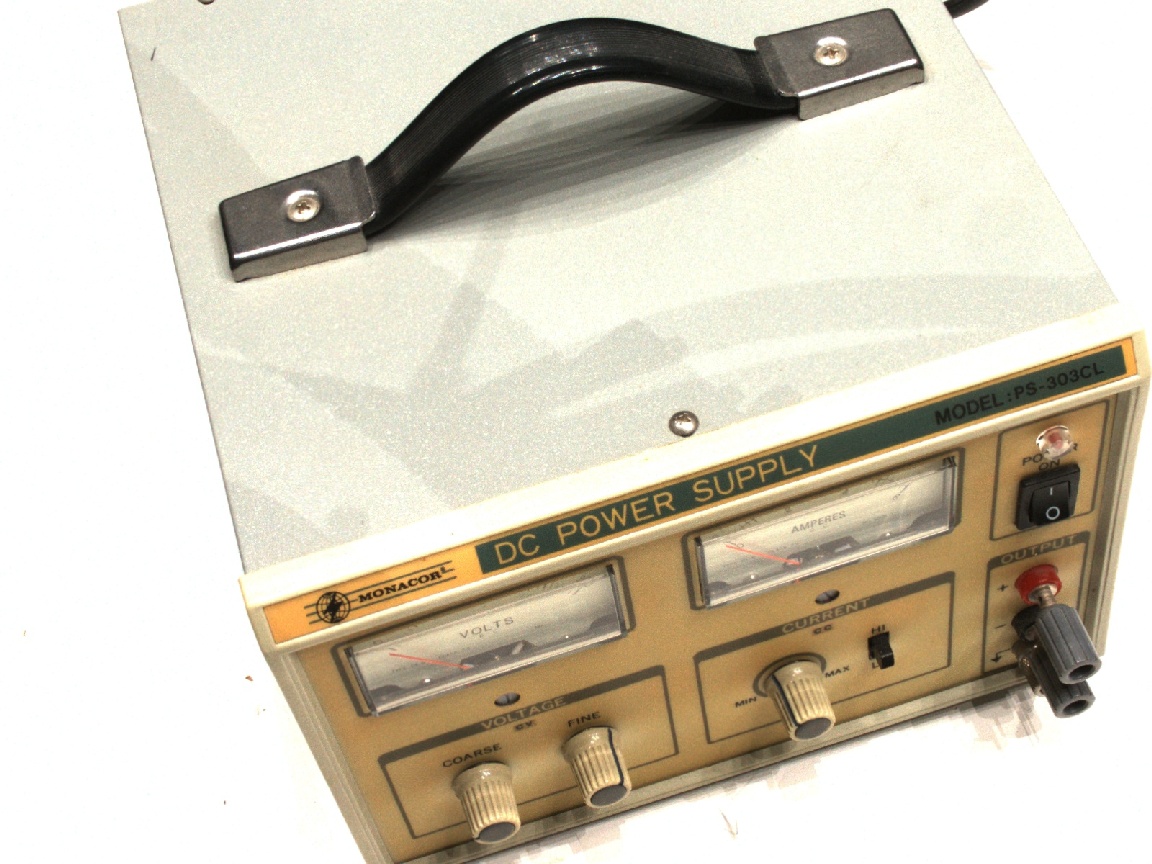} &
        \includegraphics[height=2.0cm,valign=b]{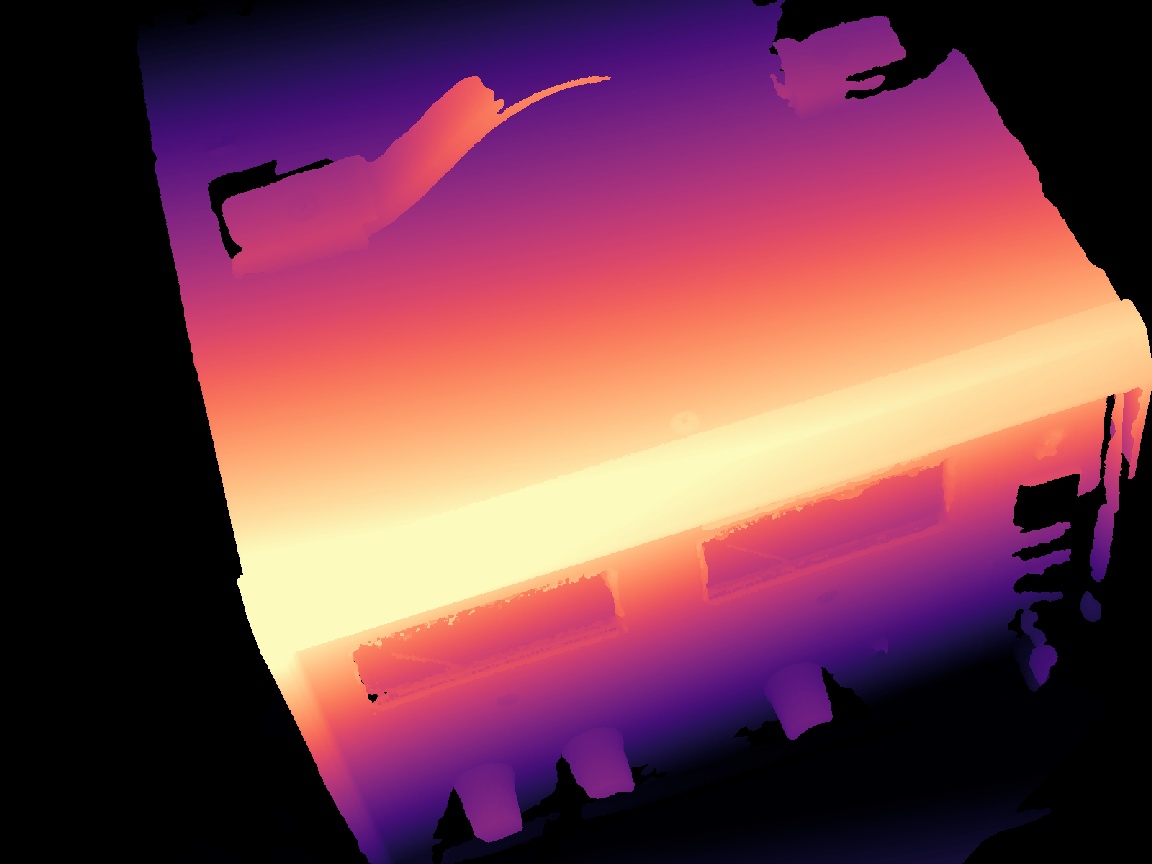} &
        \includegraphics[height=2.0cm,valign=b]{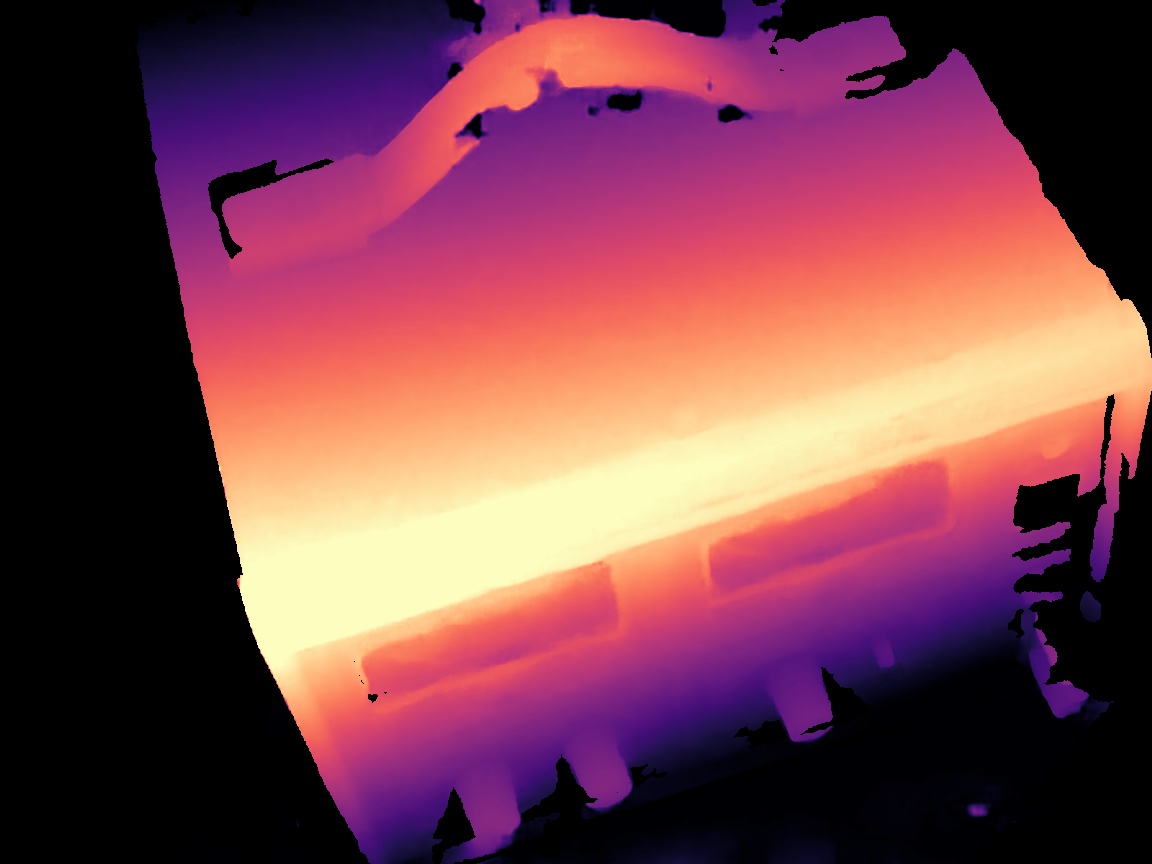} &
        \includegraphics[height=2.0cm,valign=b]{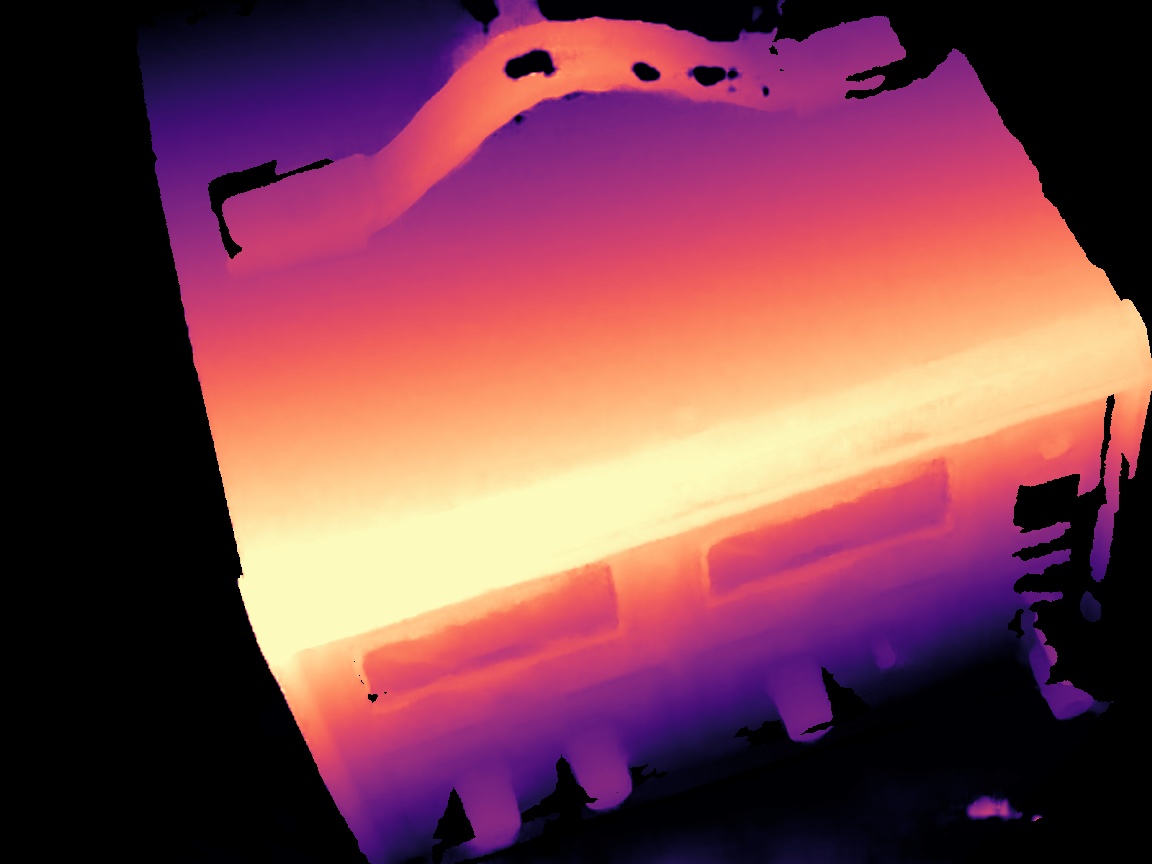} &
        \includegraphics[height=2.0cm,valign=b]{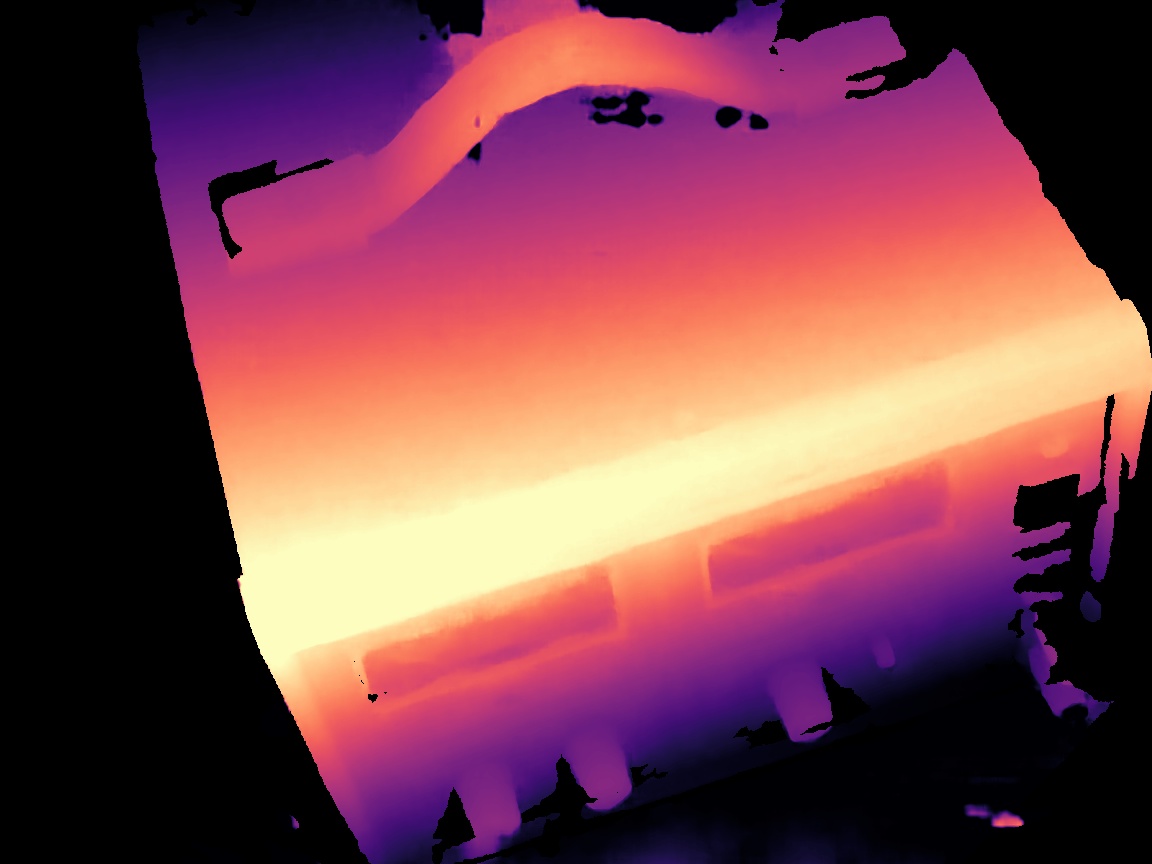} &
        \includegraphics[height=2.0cm,valign=b]{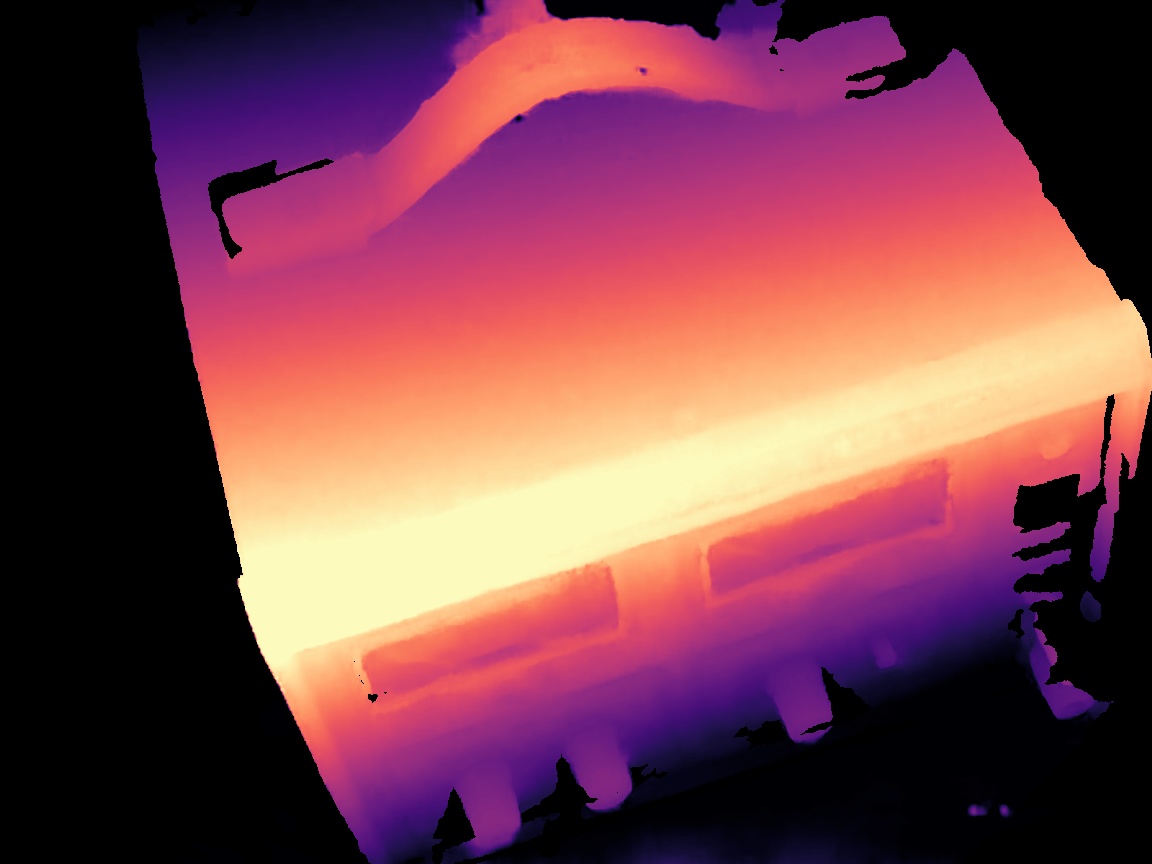} \\
\end{tabular}
\caption{Qualitative comparison of different ablation settings for CasMVSNet on the DTU dataset. From left to right: reference image, ground-truth depth, depth estimation with the full MVS-TTA framework, baseline model (CasMVSNet) without any adaptation, test-time adaptation without meta-training, and meta-auxiliary training without TTA. Our full framework produces more accurate and complete estimations, demonstrating the complementary effects of meta-training and test-time adaptation. }
\label{fig:ablation-qualitative}
\end{figure*}
\begin{figure}[t]
  \centering
  \includegraphics[width=\linewidth]{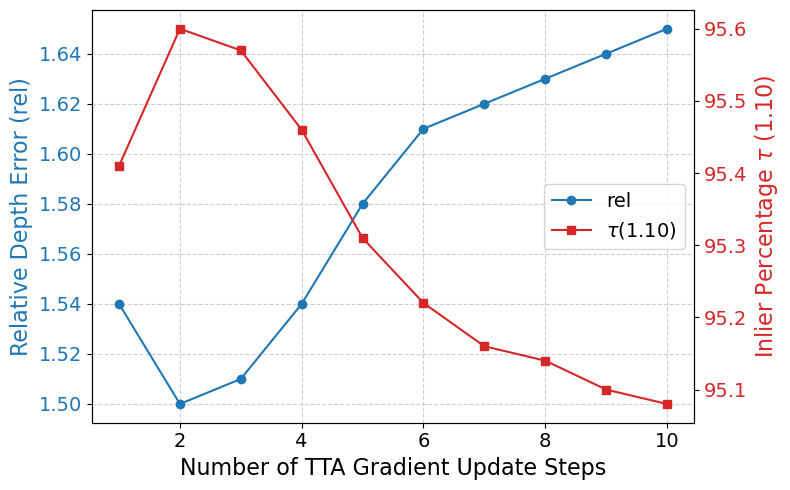}
  \caption{Effect of TTA gradient update steps on depth estimation performance on the DTU dataset using CasMVSNet as the backbone. Best results are achieved with 2 steps.}
  \label{fig:ablation-steps}
\end{figure}
We perform ablation experiments on the DTU dataset to investigate the individual contributions of test-time adaptation and meta-auxiliary learning in our framework. Results using CasMVSNet as the backbone are reported in Tab. \ref{tb:Ablation-results}, while Fig. \ref{fig:ablation-qualitative} shows a qualitative comparison. Directly applying TTA with the photometric consistency loss (without meta-training) slightly degrades performance compared to the baseline, increasing the absolute relative depth error from 1.64 to 1.83 and reducing $\tau$(1.1) from 95.1\% to 94.4\%. This indicates that without proper training, the model may overfit to noisy signals from the self-supervised loss, leading to unstable adaptation. Conversely, meta-auxiliary training alone (without applying TTA at test time) improves the absolute relative depth error to 1.75. However, the meta-trained model is not expected to perform best before adaptation but is specifically optimized to benefit from TTA. When combining meta-auxiliary learning with test-time adaptation, the full framework achieves the best results, reducing the relative depth error to 1.50 and improving $\tau$(1.03) and $\tau$(1.1) to 89.3\% and 95.6\%, respectively. These results confirm that meta-auxiliary learning effectively prepares the model for self-supervised adaptation at inference time, enabling TTA to improve depth estimation quality in a consistent and reliable manner. We further analyze the impact of the number of gradient update steps used during test-time adaptation. Fig. \ref{fig:ablation-steps} reports the absolute relative depth error and inlier percentages ($\tau$) at different step counts. We observe that performance improves with a small number of steps, reaching the best results at 2 updates. However, performance gradually deteriorates as the number of update steps increases beyond this point. This trend suggests that while lightweight adaptation is beneficial, excessive test-time adaptation may lead to overfitting to noisy photometric cues, especially in the absence of ground-truth supervision. We consistently set the update steps to 2 across all backbone models, and it yields stable performance improvements without further tuning. Finally, Tab. \ref{tb:Ablation-hyper-K} reports the sensitivity of the framework to the Top-K view selection using CasMVSNet as the backbone. Performance remains nearly invariant for $K \geq 2$ on both DTU and BlendedMVS, with only the degenerate $K = 1$ setting causing noticeable degradation as it does not follow the multi-view setting. This confirms that our method is robust to the choice of $K$ within a reasonable range.
\begin{table}[t]
  \centering
  \caption{Ablation study on the effectiveness of meta-auxiliary learning and test-time adaptation}
  \label{tb:Ablation-results}
  \begin{tabularx}{\linewidth}{l *{3}{>{\centering\arraybackslash}X}}
\toprule
                 & rel$\downarrow$             & $\tau$(1.03)$\uparrow$           & $\tau$(1.1)$\uparrow$                      \\
\midrule
CasMVSNet                           & 1.64                          & 88.9                                               & 95.1                                              \\
+ TTA (w/o meta)          & 1.83                          & 88.5                                               & 94.4                                              \\
+ Meta-Aux. (w/o TTA)     & 1.75                          & 88.1                                               & 94.4                                              \\
+ full framework          & 1.50                          & 89.3                                               & 95.6                                              \\
\bottomrule
  \end{tabularx}
\end{table}
\begin{table}[t]
\centering
\caption{Ablation study on the Top-K view selection hyperparameter $K$}
\label{tb:Ablation-hyper-K}
\begin{tabularx}{\linewidth}{l *{4}{>{\centering\arraybackslash}X}}
\toprule
    \multirow{2}{*}{Dataset} &
      \multicolumn{4}{c}{$\tau$(1.03)$\uparrow$} \\
    \cmidrule(lr){2-5}
      & $K=1$ & $K=2$ & $K=3$ & $K=4$ \\
    \midrule
BlendedMVS & 68.02 & 91.80 & 91.80 & 91.80 \\
DTU & 88.66 & 88.71 & 88.75 & 88.73 \\
\bottomrule
\end{tabularx}
\end{table}
\subsection{Comparison with Learning to Adapt for Stereo}
We compare our MVS-TTA framework with the continual test-time adaptation strategy of Tonioni et al.~\cite{tonioni2019learning}, which also builds on MAML with a self-supervised loss for stereo matching in video streams. Unlike their formulation where adaptation and supervision are performed on different frames and adaptation for each incoming frame is conditioned on all previous updates, our approach is better aligned with the MVS setting. During meta-training, our method treats each set of 5 calibrated input images as a task, so
inner adaptation using photometric consistency and outer supervision are applied to the same images. At inference time, our method performs a single TTA update from a meta-learned initialization for each test sample, rather than accumulating updates over a sequence. Tab. \ref{tb:DTU-results} shows a direct comparison using CasMVSNet with photometric consistency loss. Continuous adaptation following Tonioni et al. leads to inferior performance, likely due to a mismatch between training and testing dynamics and potential catastrophic forgetting.
\section{Conclusion}
In this paper, we propose MVS-TTA, a test-time adaptation framework that enhances the adaptability of learning-based MVS models. By combining a cross-view photometric consistency loss with meta-auxiliary learning, our method enables MVS models to self-adapt at inference time without relying on additional supervision. Extensive experiments on DTU and BlendedMVS demonstrate that MVS-TTA consistently improves performance across diverse MVS architectures. Furthermore, our method shows strong cross-dataset generalization under significant distribution shifts, confirming its robustness and scalability. MVS-TTA is model-agnostic, lightweight, and compatible with existing MVS pipelines, making it a practical extension for real-world 3D reconstruction.
\bibliographystyle{IEEEtran}
\bibliography{egbib}
\end{document}